\documentclass[pmlr]{jmlr}

\usepackage{longtable}

\usepackage{booktabs}
\usepackage[load-configurations=version-1]{siunitx} 

\theorembodyfont{\upshape}
\theoremheaderfont{\scshape}
\theorempostheader{:}
\theoremsep{\newline}

\DeclareMathOperator{\Forall}{\forall}

\DeclareMathOperator{\E}{\mathbb{E}}
\usepackage{bm}
\usepackage{bbm}
\usepackage{soul}
\usepackage{chngcntr}

\jmlrvolume{85}
\jmlryear{2019}
\jmlrworkshop{Machine Learning for Healthcare}

\title[Counterfactual Reasoning for Fair Clinical Risk Prediction]{Counterfactual Reasoning for Fair Clinical Risk Prediction}

\author{\Name{Stephen Pfohl} \Email{spfohl@stanford.edu} 
       \addr \\
       Stanford Center for Biomedical Informatics Research\\
       Stanford University \\
       Stanford, CA 
       \AND
       \Name{Tony Duan} \Email{tonyduan@stanford.edu} 
       \addr \\ 
       Department of Computer Science \\
       Stanford University\\
       Stanford, CA
       \AND
       \Name{Daisy Yi Ding} \Email{dingd@stanford.edu} 
       \addr \\
       Stanford Center for Biomedical Informatics Research \\
       Stanford University \\
       Stanford, CA       
       \AND
       \Name{Nigam H. Shah} \Email{nigam@stanford.edu} 
       \addr \\
       Stanford Center for Biomedical Informatics Research \\
       Stanford University \\
       Stanford, CA
       }
       
\begin{document}

\maketitle

\begin{abstract}
The use of machine learning systems to support decision making in healthcare raises questions as to what extent these systems may introduce or exacerbate disparities in care for historically underrepresented and mistreated groups, due to biases implicitly embedded in observational data in electronic health records. 
To address this problem in the context of clinical risk prediction models, we develop an augmented counterfactual fairness criteria to extend the group fairness criteria of equalized odds to an individual level. 
We do so by requiring that the same prediction be made for a patient, and a counterfactual patient resulting from changing a sensitive attribute, if the factual and counterfactual outcomes do not differ. 
We investigate the extent to which the augmented counterfactual fairness criteria may be applied to develop fair models for prolonged inpatient length of stay and mortality with observational electronic health records data. 
As the fairness criteria is ill-defined without knowledge of the data generating process, we use a variational autoencoder to perform counterfactual inference in the context of an assumed causal graph.
While our technique provides a means to trade off maintenance of fairness with reduction in predictive performance in the context of a learned generative model, further work is needed to assess the generality of this approach.
\end{abstract}

\section{Introduction}
The use of modern machine learning techniques capable of efficiently leveraging the full extent of the electronic health record (EHR) to make patient-specific predictions may provide the means to greatly improve the quality of care and reduce costs \citep{Goldstein2017,Bates2014,Rajkomar2018a}.
Recently, concern has been raised that the naive use of these models in routine clinical practice has the potential to reinforce existing racial, ethnic, and socioeconomic health disparities that exist in the delivery of healthcare in the United States \citep{Rajkomar2018b,Char2018,Gianfrancesco2018,Veinot2018,Cohen2014}. 
These disparities manifest both within and between healthcare institutions, and can reflect differences in patient care, differential access to healthcare resources, and differences in the incidence of conditions across groups \citep{Soto2013,Smedley2003,Mayr2010,Fowler2010,Dombrovskiy2007,Galea2007,Pines2009}.
The source of this phenomenon is complex \citep{ChenJohanssonSontag_NIPS18,Rajkomar2018b} and related to biases implicitly encoded \citep{Garg2018,kallus2018residual} in observational data through historical differences in care delivery and under-representation of minority groups in the cohorts used for model development. 

In the machine learning literature, several methods of \emph{algorithmic fairness} \citep{Chouldechova2018,Suresh2019} have been proposed. These methods provide principled approaches to reasoning about and mitigating notions of bias and discrimination in predictive models, but these tools remain under-explored in the context of clinical risk prediction. 
Recently, \citet{Rajkomar2018b} and \citet{Goodman2018} outlined a taxonomy of intended and unintended sources of bias and discrimination in healthcare along with their impact and further provided a series of practical recommendations for applying fairness constraints to machine learning models on the basis of the ethical principles appropriate to the clinical context. 
In practice, measures of fairness have been used to assess notions of inequity in the prediction of intensive care unit mortality \citep{ChenJohanssonSontag_NIPS18}, 30-day psychiatric readmission \citep{Chen2019}, risk of atherosclerotic cardiovascular disease \citep{Pfohl2018}, and for risk adjustments in health insurance markets \citep{Zink2019}. 

As of now, the prior work on the assessment of fairness for clinical predictive models has focused almost exclusively on the use of \emph{group fairness} metrics that assess a form of conditional independence between model predictions, the true outcome, and membership to a protected group on the basis of a sensitive attribute such as race, gender, or age. 
These metrics are attractive because they are straightforward to reason about and verify. However, they do not provide a meaningful assessment of fairness to individuals \citep{Dwork2012} or structured subgroups of protected demographic groups \citep{Kearns2018, Kearns2018b, pmlr-v80-hebert-johnson18a}. That is, a model that satisfies a group fairness metric may permit arbitrary discriminatory deviations from the criteria on subgroups or individuals as long as the criteria are satisfied on average across the population \citep{Kearns2018}. In contrast, \emph{counterfactual fairness} \citep{Kusner2017} is a recently proposed metric that uses tools from causal inference to assess fairness at an individual level by requiring that a sensitive attribute not be the \emph{cause} of a change in a prediction.

In this work, we provide an interpretation of fair clinical decision making from the perspective of equal benefit with respect to a sensitive attribute. We show that the group fairness notion of equalized odds \citep{Hardt2016} has a natural interpretation within this framework and argue for its use for a class of clinical prediction tasks. Furthermore, we develop an augmented counterfactual fairness formulation to extend equalized odds to the individual level.
However, since evaluation of this criteria relies on untestable causal assumptions and knowledge of the data generating process, it is generally impossible to reliably assess the extent to which a predictive model satisfies this criteria using observational data alone \citep{Pearl2009,imbens_rubin_2015,kilbertus2017avoiding}. As proof of concept, we evaluate the fairness of predictive models of inpatient mortality and prolonged length of stay using EHR data and a variational autoencoder (VAE) to perform counterfactual inference and sampling \citep{Louizos2017,Madras2018}. Within the context of counterfactual samples drawn from the VAE, we investigate the relevant trade-offs between predictive performance and fairness on our proposed metric. 

\subsection{Technical Significance}
We propose an extension of counterfactual fairness \citep{Kusner2017} and equalized odds \citep{Hardt2016} that we call individual equalized counterfactual odds. This metric is motivated by clinical risk prediction, but may be of interest to the general machine learning community for use in other applications. The algorithm we propose for developing a predictive model that satisfies this fairness metric extends counterfactual logit pairing \citep{Garg2018CF}, but relies on a VAE to simulate counterfactual samples from high dimensional and sparse EHR data. Given the practical challenges associated with the empirical evaluation of this approach, we hope that the framing we propose serves as motivation for further empirical and theoretical work at the intersection of fairness, causal inference, and deep generative models.

\subsection{Clinical Relevance}
The fairness criteria and algorithms that we propose and analyze may provide a means for interpreting and mitigating potential biases that clinical predictive models have towards historically disadvantaged groups. Our work formalizes group fairness in clinical risk prediction with a utility theoretic framework \citep{Heidari2019}. We further introduce counterfactual fairness to the clinical context as an alternative to the criteria that have been applied to clinical prediction tasks thus far.
\section{Background and Problem Formulation}
\subsection{Supervised Learning with EHR Data for Clinical Risk Prediction}
Let $X \in \mathcal{X} = \mathbb{R}^m$ be a variable designating a vector representation of coded diagnoses, procedures, medication orders, lab results, and clinical notes derived from standard EHR feature engineering or representation learning procedures \citep{Reps2018,Rajkomar2018a,Goldstein2017,Xiao2018,Miotto2016};
$Y \in \mathcal{Y} = \{0, 1\}$ be a binary indicator of the occurrence of a clinically relevant outcome;
and $A \in \mathcal{A}$ be a discrete indicator for a protected or sensitive attribute, such as race, ethnicity, gender, or age. 
We are interested  in using data $\mathcal{D} = \{(x_i, y_i, a_i)\}_{i = 1}^N \sim p(X, Y, A)$ to learn a function $h(X, A) : \mathbb{R}^{m+|\mathcal{A}|} \rightarrow [0,1]$  approximating $p(Y \mid X, A)$, which may be compared to a threshold value $T$ to produce predictions $\hat{Y}(X, A) = \mathbbm{1}[h(X, A) \geq T] \in \{0, 1\}$.

\subsection{Utility-based Clinical Motivation for Fairness}

We view the goal of fair clinical risk prediction as developing a predictive model as a component of a clinical policy that maximizes aggregate utility, while promoting health equity by requiring that the distribution of utility be independent of a sensitive attribute. This aligns with the ``equal benefit'' definition suggested by \cite{Rajkomar2018b}. A straightforward interpretation of this criteria is it requires a clinical policy to assign the same expected utility to a population partitioned by a sensitive attribute (for example, gender). Depending on the clinical context motivating algorithmic fairness, it can be appropriate to conditionally satisfy the equal benefit criteria for some collection of strata of the population $\mathcal{D}_1,\hdots,\mathcal{D}_K$ that intersect the groups defined by the sensitive attribute \citep{Heidari2019}. 

In the most general formulation, we want for strata $\mathcal{D}_1,\hdots,\mathcal{D}_K$,
\begin{equation} \label{eq:group_benefit}
    \E_{x, y \sim \mathcal{D}_k \mid A = a_i} V(h(x, a_i), y) = \E_{x, y \sim \mathcal{D}_k \mid A =a_j} V(h(x, a_j), y) \Forall a_i, a_j \in \mathcal{A}, k\in\{1,\hdots,K\}
\end{equation}
where $V$ denotes a utility function associated with a prediction $h(x,a)$ and outcome $y$. 

It is necessary to provide assumptions on the structure of policy and the individual-level utilities induced by the predictive model if we are to relate the performance characteristics of a predictive model
to a utility-based fairness notion. To that end, we assume the expected utility that a patient receives as a result of applying a predictor is given by
    \begin{align} \label{eq:utility_function}
        \begin{split}
        V(h(x, a), y) = 1 - & \alpha_0 p(\mathbbm{1}[h(x, a) \geq T] = 1 \mid Y = 0)p(Y = 0 \mid X = x,A = a) - \\ 
            & \alpha_1 p(\mathbbm{1}[h(x, a) \geq T] = 0 \mid Y = 1)p(Y = 1 \mid X = x, A = a),
        \end{split}
    \end{align}
for positive scalars $\alpha_0$ and $\alpha_1$ representing the costs of false positive and false negative errors, respectively, such that $v_i$ is bounded between 0 and 1.

These assumptions have a simple interpretation such that a perfect predictor achieves higher expected utility for each patient than any other predictor would. Furthermore, they capture the intuition that for predictive models of adverse clinical events, it is often undesirable to either over- or under- predict risk, as under-prediction of risk can lead to under-management of latent disease and result in subsequent unexpected adverse events while over-prediction of risk incurs a reduction in utility through the costs and side effects of unnecessary treatment.
We admit that this formulation is an over-simplification of clinical decision making \citep{Goodman2018}, broadly ignoring the preferences and incentives for relevant stakeholders, the limited capacity for intervention at the level of the health system, heterogeneous treatment effects, and the potential for \textit{biased labels} corrupted by historical discrimination in routine care such that more accurately predicting the label in retrospective data leads to further discrimination against the historically disadvantaged group \citep{Rajkomar2018b,kallus2018residual,Jiang2019}. However, these assumptions are often implicitly made in the on-going discussion around fairness of clinical predictive models when measures of model performance are used either as a measure of benefit or for assessing the biases of a model \citep{ChenJohanssonSontag_NIPS18,Rajkomar2018b,Chen2019,Pfohl2018}.  We believe continued work within this framework is valuable as long as these assumptions are critically evaluated regularly during the model development and deployment process.

\subsection{Group Fairness}

Among the fairness metrics frequently cited in the literature \citep{Chouldechova2018, Hardt2016, calders2009building, Zemel2013, Dwork2012,kleinberg2016inherent,Chouldechova2017}, that can be readily applied to a predictive model, the equalized odds and equality of opportunity criteria \citep{Hardt2016} are the most immediately relevant given the formulation of equation \ref{eq:group_benefit}. The equalized odds criteria is defined for a specific threshold as
\begin{equation} \label{eq:eq_odds}
    p(\hat{Y} = 1 \mid A = a_i, Y = y_k) = p(\hat{Y} = 1 \mid A = a_j, Y = y_k) \Forall a_i, a_j \in \mathcal{A}; \Forall y_k \in \mathcal{Y},
\end{equation}
and can be interpreted as requiring the same false positive rate across groups and false negative rate across groups. Crucially, the optimal predictor satisfies equalized odds \citep{Hardt2016}. Furthermore, it corresponds to the equal group benefit criteria in equation \ref{eq:group_benefit} for the utility function in equation \ref{eq:utility_function} if $\mathcal{D}_1 = \{ (x,y,a): y = 0\}$ and $\mathcal{D}_2 = \{ (x,y,a): y =1\}$. In other words, if the population is stratified on account of whether some clinical outcome $Y$ occurs or not, the same expected utility will be attained, on average, for patients drawn from groups of a sensitive attribute within the strata defined by the outcome. The formulation for equality of opportunity is similar except that data are stratified on either $Y = 0$ or $Y = 1$, but not both. 

Demographic parity \citep{calders2009building,Zemel2013} is another fairness metric that requires the probability of a positive prediction be the same for each group:
\begin{equation} \label{eq:dp}
    p(\hat{Y} = 1 \mid A = a_i) = p(\hat{Y} = 1 \mid A = a_j) \Forall a_i, a_j \in \mathcal{A}.
\end{equation}
We argue that this formulation is not desirable for clinical risk prediction tasks that follow utility function \ref{eq:utility_function} since it does not allow for the ideal predictor if the sensitive attribute is correlated with the outcome \citep{Hardt2016,Dwork2012}.

\subsection{Individual and Counterfactual Fairness}
An alternate formulation to group fairness that is, as of yet, unexplored in medicine and clinical risk prediction is that of individual fairness \citep{Dwork2012}. In general, this formulation asks that ``similar individuals be treated similarly'' \citep{Chouldechova2018},
where similarity is defined by a domain-specific metric. In contrast, group-level metrics such as equalized odds only require the criteria to hold in expectation over groups of a sensitive attribute and thus allow for arbitrary individual-level fairness deviations \citep{Kearns2018}.

The framework of counterfactual fairness \citep{Kusner2017} may be interpreted as an instance of individual fairness since it provides a means of defining a similarity metric as well as a means of assessing fairness under that metric \citep{Loftus}. 
A necessary component of this formulation is the availability of a structural equation model (SEM) \citep{Pearl2009} describing the causal relationships between latent background variables $U$ and observed variables $X,A,Y$ through a set of functional relationships that fully govern the data generating process. With an SEM, it is possible to reason about counterfactual queries such as ``what would the prediction have been for this patient if they belonged to a different group?" 

That is, we can compute $p(\hat{Y}_{A \leftarrow a'}(U) \mid X = x, A = a)$, the counterfactual distribution over $\hat{Y}$ corresponding to setting $A = a'$, given observed data $X=x, A=a$. Let $\hat{Y}_{A \leftarrow a'}(u)$ denote the value of $\hat{Y}$ obtained by computing $\hat{Y}$ for a fixed value of the background variable $U=u$ on the basis of a modified set of structural equations where $A$ is artificially set to $a'$ in all equations involving $A$. Then for observed data $X = x$ and $A = a$, counterfactual inference occurs by (1) computing the posterior $p(U|X=x, A=a)$, (2) setting $A = a'$ to create a modified set of structural equations, then (3) computing the implied distribution on $\hat{Y}$ given the modified equations and posterior on $U$. When the context is clear, we drop the argument and denote the counterfactual by $\hat{Y}_{A \leftarrow a'}$ instead of $\hat{Y}_{A \leftarrow a'}(u)$ or $\hat{Y}_{A \leftarrow a'}(U)$.

A predictor $\hat{Y}$ is then \emph{counterfactually fair} if for any $x \in \mathcal{X}$ and $\Forall y \in \mathcal{Y}, a, a' \in \mathcal{A}$:
\begin{equation} \label{eq:cf_fairness}
    p(\hat{Y}_{A \leftarrow a}(U) = y \mid X = x, A = a) = p(\hat{Y}_{A \leftarrow a'}(U) = y \mid X = x, A = a).
\end{equation}
The interpretation of this criterion is that it requires the same distribution of predictions for each individual in the factual world where $A = a$ and in each counterfactual world where $A = a'$, for all $a' \neq a \in \mathcal{A}$. As such, it disallows a sensitive attribute to be the \textit{cause} of a change in the prediction.
The individual-level distance metric that is implied by this formulation is that individuals are treated as close to a set of matched counterfactual individuals that share the same value for the background variables $U$ but differ in their membership to a group of a sensitive attribute \citep{Loftus}.

\section{Counterfactual Reasoning for Fair Clinical Risk Prediction}

\subsection{Individual Equalized Counterfactual Odds}

We now propose a new criterion \textbf{individual equalized counterfactual odds}, which is satisfied if 
for all $x \in \mathcal{X}, a, a' \in \mathcal{A}, y \in \mathcal{Y}$:
\begin{multline} \label{eq:ind_eq_odds}
   p(\hat{Y}_{A \leftarrow a}(U) \mid X = x, Y_{A \leftarrow a} = y, A = a) = p(\hat{Y}_{A \leftarrow a'}(U) \mid X = x, Y_{A \leftarrow a'} = y, A = a).
\end{multline}
This ensures the predictor is counterfactually fair, \emph{conditioned on the factual outcome $Y$ matching the counterfactual outcome $Y_{A \leftarrow a'}$}. In contrast, the original counterfactual fairness formulation can be interpreted as requiring predictions to be the same across factual-counterfactual pairs, regardless of whether those pairs share the same value of the outcome.

To connect to our utility-based motivation for fairness, a natural desiderata is one where the clinical policy assigns the same expected utility to each individual patient in the factual world and in expectation over the set of counterfactual worlds where the sensitive attribute is set to some other value. In other words, the individual's sensitive attribute should not be the \textit{cause} of a reduction in utility relative to the utility they would receive if they belonged to some other group of a sensitive attribute. Counterfactual fairness only implies this property if we assume that the utility function does not depend on the outcome $Y$ and positive predictions are unambiguously preferred.
Individual equalized counterfactual odds may then be interpreted as requiring that this individual utility criteria hold conditioned on holding $Y$ constant for the factual and counterfactual individual, thus providing an individual-level counterfactual analog to equalized odds. 

Overall, we are then interested in building predictive models that:
\begin{enumerate}
    \item For samples drawn from the factual distribution $\{x, y, a\}$, predict $y$ as well as possible.
    \item For the counterfactual samples $\{x_{A \leftarrow a'}, y_{A \leftarrow a'}, a'\}$ predict $y_{A \leftarrow a'}$ as well as possible.
    \item Satisfy individual equalized counterfactual odds.
\end{enumerate}

\subsection{Training a Fair Predictor}
Now, we provide a practical training objective that can be used to develop a predictor that satisfies the proposed criteria. We assume access to a SEM that may be used for sampling counterfactuals with respect to a sensitive attribute. Let $h_{\theta}$ be a black-box predictor, such as a neural network, with parameters $\theta$; $J(h_{\theta}(x, a), y)$ be the cross-entropy loss; and $\sigma$ corresponds to the sigmoid function. The loss $\mathcal{L}$ for a sample $\{x, y, a\} \sim p(X, Y, A)$ is as follows:
\begin{align} \label{eq:clp_loss}
    \begin{split}
     \mathcal{L} & = J(h_{\theta}(x, a), y) +
     \lambda_\mathrm{CF} \sum_{a_k \in\mathcal{A}} \mathbbm{1}[a \neq a_k] J(h_{\theta}(x_{A \leftarrow a_k}, a_k), y_{A \leftarrow a_k}) + \\
     & \lambda_\mathrm{CLP} \sum_{a_k \in\mathcal{A}} \mathbbm{1}[a \neq a_k]\mathbbm{1}[y = y_{A \leftarrow a_k}] \Big(\sigma^{-1}(h_{\theta}(x_{A \leftarrow a_k}, a_k)) - \sigma^{-1}(h_{\theta}(x, a))\Big)^2
     \end{split}
\end{align}
where $\lambda_\mathrm{CF}$ and $\lambda_\mathrm{CLP}$ are scalar hyperparameters that may be used to control the relative contribution of the three components of the loss. The first term corresponds to the loss incurred due to errors the predictor makes on the factual sample, the second term to the loss on the counterfactual sample, and the third term is a counterfactual logit pairing (CLP) term \citep{Garg2018CF,Kannan2018}, which is used to encourage the model to satisfy individual equalized counterfactual odds. 

\begin{figure}
	\centering
	\includegraphics[width=0.21\linewidth]{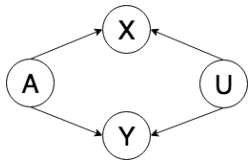}
	\caption{Structure of the assumed causal model. Unobserved latent variables $U$ and sensitive attribute $A$ jointly generate the observed data $X$ and the outcome $Y$.}
	\label{fig:causal_model}
\end{figure}

\subsection{Causal Effect VAE for Counterfactual Inference}
Our training objective requires an SEM for performing counterfactual inference with respect to a sensitive attribute. However, an SEM accurately describing the causal relationships among unobserved confounders, sensitive demographic attributes, relevant clinical outcomes, and the high-dimensional set of covariates extracted from the EHR is rarely readily available in practice. Without additional assumptions, it is generally impossible to infer the causal structure of the underlying data generating process directly from the observable properties of an observational dataset \citep{Pearl2009,imbens_rubin_2015}.

In practice, we employ a causal effect VAE \citep{Louizos2017} to model causal effects in the presence of unobserved confounders with observable proxies \citep{Kuroki2014,Miao2018,Madras2018,Wang2018,wang2019multiple,Tran2017}. Previously, \cite{Louizos2017} and \cite{Madras2018} established a sufficient condition for causal identifiability in a related setting by requiring a well-specified causal model (i.e. a directed acyclic graph indicating the presence and directionality of causal relationships between variables with appropriate prior distributions on unobserved variables) and an assumption that it is possible to estimate the true joint distribution $p(X,Y,A,U)$ from a finite sample observational dataset drawn from $p(X, Y, A)$. However, as the causal graph and parameters of interest that we consider differ from those considered in these prior works, we make no formal guarantee of identification even in the case where these assumptions hold.

In our experiments, we assume a causal graph similar to that used in \cite{Madras2018}, depicted in Figure \ref{fig:causal_model}, such that $X, Y$ are causally downstream of $A$ and unobserved confounders $U$, and that $X$ and $Y$ are independent conditioned on $U$ and $A$. 

The assumed generative process is as follows: $u$ is drawn from an isotropic Gaussian prior, $a$ is drawn from a multinomial distribution with marginals $\pi$, and $x$ and $y$ are drawn from complex distributions, but are independent given $a$ and $u$. 
\begin{align*}
    u & \sim p(U)  = \textrm{Normal}(0, I) \\
    a &\sim p(A)  = \textrm{Categorical}(A \mid \pi) \\
    x, y &\sim p(X, Y \mid U, A) = p(X \mid U, A)p(Y \mid U, A).
\end{align*}
We introduce parameterized functions $p_{\theta}(x \mid u, a)$, $p_{\theta}(y \mid u, a)$, and $q_{\phi}(u \mid x, a)$ 
and learn parameters $\theta,\phi$ via the following loss function that when minimized, maximizes the lower bound on the marginal log-likelihood \citep{Kingma2013a,Rezende2014}:
\begin{equation}
    \mathcal{L}_{\text{ELBO}} = -\mathbb{E}_{u \sim q_{\phi}(u\mid x, a)}[\log p_{\theta}(x \mid u, a) + \log p_{\theta}(y \mid u, a)] + D_\mathrm{KL}(q_{\phi}(u \mid x,a) \mid \mid p(u)).
\end{equation}

As this objective is known to permit degenerate solutions where the latent variables contain no mutual information with the observed data \citep{Bowman2016,Alemi2018,ZhaoInfo,He2019}, we employ a variant of the InfoVAE objective \citep{ZhaoInfo,Tolstikhin2017} that instead of directly maximizing the ELBO, leverages a divergence over the \textit{aggregated posterior} $q_\phi(u)$, implicitly regularizing against a loss of mutual information \citep{ZhaoInfo,kim2018disentangling},
\begin{equation} \label{eq:InfoVAE}
    \mathcal{L}_{\text{InfoVAE}} = -\mathbb{E}_{u \sim q_{\phi}(u \mid x, a)}[\log p_{\theta}(x \mid u, a) + \log p_{\theta}(y \mid u, a)] + \lambda D(q_{\phi}(u) \mid\mid p(u))
\end{equation}
where $D$ is any divergence and $\lambda$ is a positive scalar. Here, we choose $D$ to be the Maximum Mean Discrepancy (MMD) \citep{gretton2012kernel} with a Gaussian radial basis function kernel, due to the robust empirical performance of this metric in \cite{ZhaoInfo}. 

We apply an additional constraint to the loss function to encourage the approximate posterior $q_{\phi}(u \mid a)$ to be independent of $a$, similar to \cite{Louizos2015} and \cite{Chiappa2019}. 
Overall, for a training set $\mathcal{D}$ we minimize a weighted version of the loss,
\begin{multline} \label{eq:loss_VAE}
     \mathcal{L}_\mathrm{CE-VAE} = \mathbb{E}_{(x,y,a) \sim \mathcal{D}}\left[ -\mathbb{E}_{u \sim q_{\phi}(u \mid x, a)}\left[\lambda_x \log p_{\theta}(x \mid u, a) + \lambda_y \log p_{\theta}(y \mid u, a)\right]\right] + \\ \lambda_\mathrm{MMD} D_\mathrm{MMD}(q_{\phi}(u) \mid\mid p(u)) + \lambda_{\mathrm{MMD}_A} \sum_{a_k\in\mathcal{A}} D_\mathrm{MMD}(q_{\phi}(u \mid a = a_k) \mid\mid p(u)).
\end{multline}
where $\lambda_x$, $\lambda_{y}$, $\lambda_\mathrm{MMD}$, $\lambda_{\mathrm{MMD}_A}$ are scalar hyperparameters. 

\section{Methods}
\begin{table}[tb]
\footnotesize
\centering 
    \begin{tabular}{lrrr}
    \toprule
        Group &   Count &  Length of Stay $ \geq 7$ Days &  Inpatient Mortality \\
    \midrule
        Asian &   17,465 &         0.187 &               0.025 \\
        Black &    5,202 &         0.239 &               0.020 \\
     Hispanic &   21,978 &         0.196 &               0.019 \\
        Other &   11,004 &         0.200 &               0.022 \\
      Unknown &    3,593 &         0.201 &               0.072 \\ 
        White &   70,391 &         0.204 &               0.021 \\ \midrule
       Female &   72,556 &         0.167 &               0.018 \\
         Male &   57,076 &         0.245 &               0.029 \\ \midrule
     \lbrack 18, 30) &   15,291 &         0.180 &               0.007 \\
     \lbrack 30, 45) &   27,155 &         0.140 &               0.007 \\
     \lbrack 45, 65) &   43,529 &         0.222 &               0.025 \\
     \lbrack 65, 89) &   43,658 &         0.226 &               0.036 \\ \midrule
          All &  129,633 &         0.201 &               0.023 \\
    \bottomrule
    \end{tabular}
    \caption{Cohort characteristics. Shown are the number of patients and incidence of prolonged length of stay and inpatient mortality for each race/ethnicity, gender, and age group.} \label{tab:cohort_characteristics}
\end{table}

\subsection{Cohort Construction and Labeling}
We extract records from the Stanford Medicine Research Data Repository \citep{Lowe2009}, a clinical data warehouse containing records on roughly three million patients for clinical encounters occurring between 1990 and 2018.
We extract all inpatient admissions for patients eighteen years or older that occur in January 2010 or later with a duration longer than 24 hours and assign an index time at 11:59 PM on the night of admission.
If a patient has more than one valid admission meeting this criteria, we randomly select one for entry into the cohort. We consider two outcomes: (1) inpatient mortality, defined as death prior to discharge from the hospital, and (2) prolonged length of stay (LOS), defined as a stay lasting seven days or longer. We consider three sensitive attributes in our experiments: (1) race/ethnicity, defined as Hispanic if ethnicity is recorded as Hispanic and the value of the recorded race otherwise, (2) gender, recorded as male or female\footnote{Only one patient meeting the cohort inclusion criteria was recorded with a label other than male or female. We exclude that patient for experiments for which gender is a sensitive attribute of interest, but include them otherwise.}, and (3) age at the index time, discretized into four disjoint groups: 18-29, 30-44, 45-64, and 65-89 years of age.

Statistics describing the relevant counts of patients per group and incidence of inpatient mortality and prolonged length of stay for each group are displayed in Table \ref{tab:cohort_characteristics}. Notably, of 129,633 unique patients in the final cohort, 70,391 of them are labeled as of white race, constituting a majority; the black population has an elevated incidence of prolonged length of stay relative to other groups; the incidence of prolonged length of stay and inpatient mortality appears to increase with age; and a small population of patients labeled with unknown race experience an elevated mortality rate. For the purposes of model development and evaluation, the patients are randomly partitioned such that 80\%, 10\%, 10\% are used for training, validation, and testing, respectively. 

\subsection{Feature Extraction and Representation}
To construct a feature representation suitable for prediction, we begin by filtering the historical record for each patient for those recorded prior to the index time. We construct a dictionary of unique clinical concepts over the set of filtered patient records by mapping each unique historical diagnosis, procedure order, prescription, lab test order, and encounter type to a unique token in the dictionary\footnote{We consider only coded clinical concepts stored in their source vocabularies, do not extract information from clinical notes, and do not leverage numeric data such as vitals or the results of lab tests.}. We then construct a sparse binary feature representation for each patient by encoding each element of the dictionary as a binary attribute indicating whether that element occurred at any point in the historical record for the patient prior to the index time. When training models to be fair with respect to a sensitive attribute, we remove that sensitive attribute from the feature space and append all other demographic variables not considered sensitive. The dictionary size is 368,117 features, including all demographic variables.

\subsection{Modeling and Evaluation}
We conduct a series of experiments that aim to assess the practical capability for and implication of developing clinical risk prediction models that satisfy the individual equalized counterfactual odds criteria, when counterfactuals are sampled from a causal effect VAE approximating the appropriate SEM. 
Experiments are replicated separately for each of the three sensitive attributes (race/ethnicity, gender, and age) and the two clinical outcomes (length of stay and inpatient mortality), for six experiments total. As a baseline, we train a fully-connected feedforward neural network to predict each outcome using all features and sensitive attributes as input (details in Appendix \ref{appendix:training}). 

For each combination of sensitive attribute and clinical outcome, we first train a causal effect VAE to approximate the corresponding SEM (details in Appendix \ref{appendix:training}). We then train a classifier $h_{\theta}$ to be fair with respect to the approximated SEM, by minimizing the loss in Equation \ref{eq:clp_loss} with samples drawn from the corresponding VAE (details in Appendix \ref{appendix:training}).
We perform a grid search over a set of hyperparameters that includes a logarithmic scale over $\lambda_\mathrm{CLP}$, $\lambda_\mathrm{CF}$, and learning rates (details in Appendix \ref{appendix:params}).
When reporting results, we select among these models the one that minimizes the unweighted CLP component of Equation \ref{eq:clp_loss} on the validation set for each unique value of $\lambda_\mathrm{CLP}$.
All models were developed using the Pytorch framework \citep{paszke2017automatic}.

For all models, we evaluate the area under the Receiver Operating Characteristic curve (AUC-ROC), the area under the Precision-Recall curve (AUC-PRC), and the Brier score on the full test set as well as on the subgroups corresponding to the sensitive attribute of interest. Additionally, for each patient in the test set, we compute the predicted probability of the outcome produced by the predictor and compute the difference between the counterfactual and factual predictions in a pairwise fashion. When conditioning on the value of the outcome across these factual-counterfactual pairs, we obtain a measure of individual equalized counterfactual odds for each factual-counterfactual pair for each individual. We use the mean difference for the factual-counterfactual transition across a population, conditioned on the outcome, to assess the bias a predictor has for one group versus another.

\section{Results}

\begin{table}[tb]
    \centering 
    \footnotesize
    \begin{tabular}{lrrrrrrrr}
    \toprule
    {} & \multicolumn{4}{c}{Length of stay $\geq 7$ Days} & \multicolumn{4}{c}{Inpatient Mortality} \\
    \cmidrule(lr){2-5} \cmidrule(lr){6-9}
    $\lambda_\mathrm{CLP}$ & AUC-PRC & AUC-ROC & Brier &      CLP &   AUC-PRC & AUC-ROC &  Brier &      CLP \\
    \midrule
    N/A &   0.582 &   0.851 & 0.115 &      N/A &     0.267 &   0.893 & 0.0206 &      N/A \\
    0.0      &    0.56 &   0.843 &  0.12 &   0.0237 &     0.193 &   0.859 & 0.0254 &   0.0929 \\
    0.01     &   0.564 &   0.844 & 0.117 &   0.0106 &     0.192 &   0.856 &  0.025 &   0.0189 \\
    0.1      &   0.563 &   0.844 & 0.117 &  0.00111 &     0.186 &    0.82 & 0.0253 &  0.00456 \\
    1.0      &   0.563 &   0.845 & 0.117 & 0.000111 &     0.197 &     0.8 & 0.0228 & 0.000355 \\
    10.0     &   0.563 &   0.843 &  0.12 & 2.44e-05 &     0.194 &     0.8 & 0.0241 & 1.22e-06 \\
    \bottomrule
    \end{tabular}
    \caption{Model performance as a function of $\lambda_{\mathrm{CLP}}$ when race/ethnicity is considered as a sensitive attribute. CLP is an aggregate measure of the extent to which a model satisfies individual equalized counterfactual odds and is computed as the mean per factual sample of the third term in equation \ref{eq:clp_loss}. N/A indicates the baseline model.} \label{tab:race_performance}
\end{table}

\begin{table}[tb]
    \centering 
    \footnotesize
    \begin{tabular}{llrrrrrr}
    \toprule
          \multicolumn{2}{c}{}   &    \multicolumn{6}{c}{$\lambda_\mathrm{CLP}$}   \\
          \cmidrule(lr){3-8}
    Group & Metric &           N/A &    0.0 &   0.01 &    0.1 &    1.0 &   10.0        \\
    \midrule
    Asian & AUC-PRC &     0.605 & 0.563 & 0.555 & 0.561 &  0.56 & 0.562 \\
          & AUC-ROC &      0.86 & 0.853 & 0.853 & 0.854 & 0.849 & 0.851 \\
          & Brier &     0.106 &  0.11 & 0.109 & 0.109 &  0.11 & 0.112 \\
    Black & AUC-PRC &     0.579 & 0.548 &  0.55 & 0.545 & 0.563 & 0.573 \\
          & AUC-ROC &     0.838 & 0.825 &  0.82 & 0.825 & 0.823 & 0.823 \\
          & Brier &     0.124 & 0.135 & 0.129 & 0.128 & 0.127 & 0.129 \\
    Hispanic & AUC-PRC &     0.592 & 0.558 & 0.565 &  0.57 & 0.564 &  0.56 \\
          & AUC-ROC &     0.862 & 0.855 & 0.856 & 0.861 & 0.853 & 0.854 \\
          & Brier &     0.113 & 0.117 & 0.115 & 0.114 & 0.117 & 0.118 \\
    Other & AUC-PRC &     0.549 & 0.557 & 0.557 & 0.563 & 0.553 & 0.561 \\
          & AUC-ROC &     0.824 & 0.827 & 0.819 & 0.824 & 0.819 & 0.827 \\
          & Brier &     0.122 & 0.124 & 0.121 & 0.121 & 0.122 & 0.124 \\
    Unknown & AUC-PRC &     0.675 & 0.616 & 0.616 & 0.606 & 0.614 & 0.633 \\
          & AUC-ROC &       0.9 & 0.891 & 0.888 & 0.893 & 0.891 & 0.887 \\
          & Brier &     0.104 & 0.106 & 0.103 & 0.103 & 0.105 & 0.111 \\
    White & AUC-PRC &     0.575 & 0.568 & 0.564 & 0.559 & 0.562 & 0.563 \\
          & AUC-ROC &     0.847 &  0.84 & 0.839 & 0.838 & 0.838 & 0.837 \\
          & Brier &     0.118 &  0.12 & 0.118 &  0.12 &  0.12 & 0.121 \\
    \bottomrule
    \end{tabular}
    \caption{Model performance for prediction of prolonged length of stay on each group as a function of $\lambda_{\mathrm{CLP}}$ when race/ethnicity is considered as a sensitive attribute. N/A indicates the baseline model.} \label{tab:los_race_performance}
\end{table}

For each combination of sensitive attribute (race/ethnicity, gender, age) and clinical outcome (length of stay, inpatient mortality), we train a series of predictive models that are penalized, to varying degrees, against individual-level deviations in the prediction logit on the factual samples versus counterfactual samples that share the same value of the clinical outcome. The counterfactuals are obtained on the basis of an intervention on a sensitive attribute in a causal effect VAE trained to approximate the data generating process.

In the interest of brevity, in the main text, we present results corresponding to the prolonged length of stay outcome with race/ethnicity treated as the sensitive attribute. Results for other sensitive attributes (age, gender) as well as for combinations of those attributes with the mortality outcome are provided in Appendix \ref{appendix:tables} and \ref{appendix:figures}. 
\subsection{Baseline Model Performance}
In this section, we discuss aggregate performance of the baseline model (a feed-forward neural network including all sensitive attributes). We will later compare to these results when discussing the models developed to satisfy individual equalized counterfactual odds. 

For prolonged length of stay the baseline model attains an AUC-PRC of 0.582, an AUC-ROC of 0.851, and a Brier score of 0.115 (Table \ref{tab:race_performance}). For inpatient mortality the baseline model attains an AUC-PRC of 0.267, an AUC-ROC of 0.893, and a Brier score of 0.0206. We note that the model exhibits disparate performance across subgroups defined by each sensitive attribute. Across subgroups defined by race/ethnicity, the model for prolonged length of stay generally performs comparably across the subgroups, with the worst performance on the group labeled as ``Other'' (AUC-PRC of 0.549 and AUC-ROC of 0.862) and the best performance for the group labeled as ``Unknown'' (AUC-PRC of 0.675 and AUC-ROC of 0.90). Across subgroups defined by gender, the model for prolonged length of stay exhibits lower AUC-ROC and worse calibration for the male population (Table \ref{tab:los_gender_performance}). Across subgroups defined by age, the model for prolonged length of stay exhibits lower AUC-ROC and worse calibration as age increases (Table \ref{tab:los_age_performance}). For the model of inpatient mortality, results are more variable across groups (Table \ref{tab:mortality_race_performance}, \ref{tab:mortality_gender_performance}, \ref{tab:mortality_age_performance}), likely due to the lower prevalence of positive labels.

\begin{figure}[tb]
	\centering
	\includegraphics[width=0.99\linewidth]{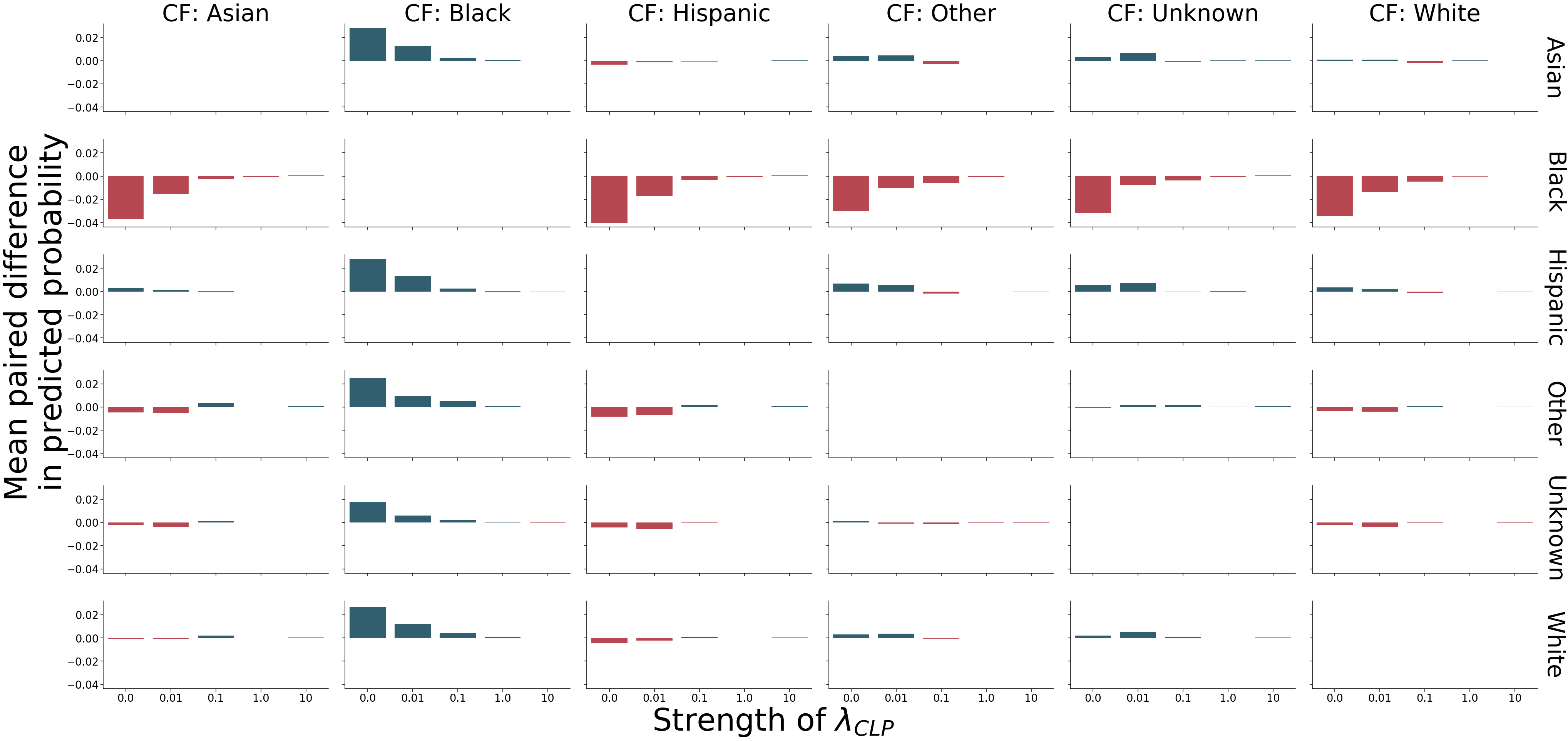}
	\caption{Mean difference in the counterfactual versus factual predicted probability of a length of stay greater than or equal to seven days conditioned on the outcome \textbf{not occurring} across race/ethnicity factual-counterfactual pairs. Positive values indicate a larger value for the counterfactual relative to the factual prediction.}
	\label{fig:bar_plot_los_race_0}
\end{figure}

\begin{figure}[tb]
	\centering
	\includegraphics[width=0.99\linewidth]{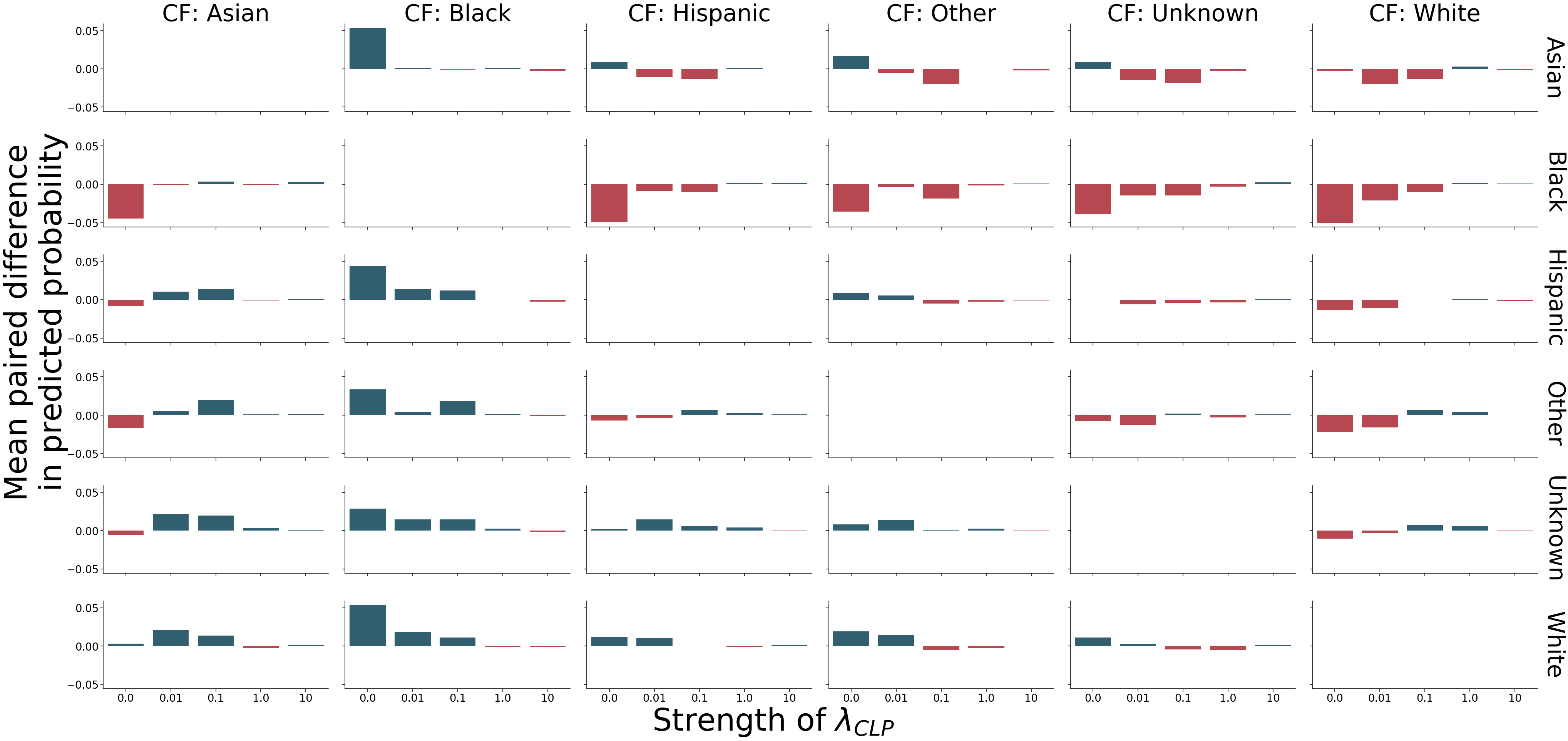}
    \caption{Mean difference in the counterfactual versus factual predicted probability of a length of stay greater than or equal to seven days conditioned on the outcome \textbf{occurring} across race/ethnicity factual-counterfactual pairs. Positive values indicate a larger value for the counterfactual relative to the factual prediction.}
	\label{fig:bar_plot_los_race_1}
\end{figure}

\subsection{Trade-offs in Fairness}
For models of prolonged length of stay, we observe that the aggregate model performance does not appear to degrade as a function of  $\lambda_\mathrm{CLP}$, although there does appear to be a fixed minor reduction in AUC-ROC and AUC-PRC for each of these models relative to the baseline (Table \ref{tab:los_race_performance}). However, for models that predict inpatient mortality, we observe more significant trade-offs in terms of reduced AUC-ROC and AUC-PRC at higher levels of $\lambda_\mathrm{CLP}$ (Table \ref{tab:mortality_race_performance}). These trends are generally reproduced for each subgroup in terms of the relative changes in the aggregate performance measures for each subgroup over the range of $\lambda_\mathrm{CLP}$ (Table \ref{tab:los_race_performance}, \ref{tab:mortality_race_performance}) relative to the subgroup-level baseline.

When the mean difference between counterfactual and factual predictions conditioned on an equal factual-counterfactual outcome are computed in a pairwise fashion between all pairs of groups, we attain an aggregate measure indicating to what extent a predictor satisfies individual equalized counterfactual odds. Furthermore, the directionality and magnitude of these differences for models trained with $\lambda_\mathrm{CLP} = 0$ gives a measure of bias that a predictive model developed with standard procedures may have towards a group.
For instance, in the case that $\lambda_\mathrm{CLP} = 0$, we observe that the mean predicted probability of a prolonged length of stay is reduced for black patients for transitions to any other counterfactual group, conditioned on the outcome not changing in the counterfactual (Figures \ref{fig:bar_plot_los_race_0}, \ref{fig:bar_plot_los_race_1}). We see that the opposite is true as well in that, on average, counterfactual transitions from any group towards a black counterfactual race/ethnicity leads to an increased prediction of prolonged length of stay conditioned on the length of stay being the same for the factual-counterfactual pair. In other experiments, the inpatient mortality prediction model shows qualitatively similar behavior on the ``Unknown" race/ethnicity group (Figures \ref{fig:bar_plot_mortality_race_0}, \ref{fig:bar_plot_mortality_race_1}). We find that our approach appears to be capable of mitigating these differences, as the relative magnitude of these differences greatly reduces for modest values of $\lambda_\mathrm{CLP}$.

\section{Discussion}
In this work, we develop an individual-level analogue to the equalized odds criterion using the counterfactual fairness framework and provide a practical algorithm applied to EHR data that leverages a causal effect VAE to perform counterfactual inference. We empirically evaluate this approach and show that we can produce predictive models of prolonged length of stay that achieve fairness with respect to individual equalized counterfactual odds, in the context of the learned generative model, with only minor reductions in aggregate performance metrics. However, we find that these trade-offs are more severe for models that predict inpatient mortality.

\subsection{Related Causally-Motivated Fairness Frameworks}
Our formulation of individual equalized counterfactual odds is connected to several other causally-motivated techniques for measuring fairness. The most related is the metric proposed in \cite{AcovRitov}, which bears similarity to equation $\ref{eq:ind_eq_odds}$ except that it does not condition on $X = x$. Our work is also related to a series of works that define \emph{path-specific} measures of fairness \citep{Nabi2018a,Chiappa2019,kilbertus2017avoiding}, which can be interpreted as a relaxation of counterfactual fairness where fairness is only enforced through pre-defined paths in a causal graph. \cite{Zhang2018a} explore the use of graphical causal explanation techniques to provide an alternative notion of a causal analogue to equalized odds and equality of opportunity that allows for potential sources of bias to be decomposed into interpretable components. Finally, the formulation of the VAE we use for counterfactual inference is inspired by the work of \cite{Madras2018}, who use a similar causal diagram and model architecture to estimate heterogeneous treatment effects by treating sensitive attributes as causally upstream of observed data.

\subsection{Fairness and Utility}
While we motivate our work on the basis of clinical contexts for which individual- and group- notions of equalized odds are appropriate measures of fairness due to a correspondence between these measures to equitable utility maximization, this formulation is only appropriate under the strong assumptions that we place on the structure of the utility function and policy implied by the predictive model. It should be emphasized that other commonly cited measures of group fairness, including demographic parity \citep{calders2009building,Zemel2013,Dwork2012} and predictive value parity \citep{Chouldechova2017,kleinberg2016inherent,Heidari2019} can also be cast in the equal benefit framework we describe here if appropriate conditioning sets and a utility function consistent with the clinical context are specified. For instance, if it is assumed that all patients prefer a positive prediction and the utility function does not depend on the outcome $Y$, then demographic parity \citep{calders2009building, Zemel2013} is appropriate. Furthermore, in cases where clinicians have limited capacity to intervene and dismissal bias or alert fatigue are a concern \citep{Rajkomar2018b}, it may be more appropriate to design fairness criteria around equalizing the positive and negative predictive values or calibration across groups \citep{kleinberg2016inherent,Chouldechova2017,Heidari2019}.

\subsection{Causal Identifiability and the VAE}
A limitation of our approach is the reliance on a VAE to perform counterfactual inference for observational datasets when an SEM is not available, as it is generally impossible to verify the assumptions sufficient for identification of the relevant causal effects \citep{Louizos2017,kilbertus2017avoiding,Madras2018}. 
Without identification, a counterfactual fairness criteria that relies on a learned generative model for counterfactual sampling, including the original formulation of \cite{Kusner2017} and individual equalized counterfactual odds, is only well-defined in the context of the generative model, which may differ arbitrarily from the true data generating process as long as the corresponding observational distributions match. 
It is thus possible that a predictive model deemed fair on the basis of individual equalized counterfactual odds may in truth be unfair with respect to the true data generating process.
In future work, we plan to investigate the use of sensitivity analyses \citep{Franks2018,DAmour2019} and simulation studies to explore the effect that these considerations have on the the procedure we propose.

\section{Conclusion}
We build off of recent efforts to formalize notions of algorithmic fairness for clinical decision support systems based on predictive models. In doing so, we propose a novel counterfactual fairness measure called individual equalized counterfactual odds and argue for its use for a class of clinical risk prediction problems where it is of interest to produce accurate predictive models that are fair to individuals. Empirically, the training procedure we propose is capable of producing fair clinical risk prediction models from EHR data with respect to individual equalized counterfactual odds computed on the basis of counterfactuals sampled from a learned generative model, but further work is needed to characterize the robustness and consistency of our approach. 

\bibliography{references}
\newpage
\appendix
\counterwithin{figure}{section}
\counterwithin{table}{section}
\section{Additional Training Details} \label{appendix:training}
\subsection{Baseline Predictor}
We construct a fully-connected neural network with fixed size per layer as a baseline predictor for each outcome. We consider the number of layers, the size of each layer, the learning rate, the dropout probability, and the use of layer normalization as hyperparameters. For each outcome, we select the model that minimizes the validation loss over one hundred iterations of random search. The selected hyperparameters are shown in Table \ref{tab:params_baseline}.

\subsection{Causal Effect VAE}
For each sample $\{x, y, a\}$, the amortized inference model $q_{\phi}(u \mid x, a)$ maps from the sparse and high dimensional input to the parameters of a Gaussian $\{\mu, \sigma\}$ with diagonal covariance and the reparameterization trick \citep{Kingma2013a} is used to draw a single sample $u$. In the latent space, we embed and concatenate $a$ to $u$ as input to the decoders $p_{\theta}(y \mid u, a)$ and $p_{\theta}(x \mid u, a)$. As the data $X$ are binary and we model $p_{\theta}(x \mid u, a)$ with components conditionally independent given $u$ and $a$, we perform maximum likelihood in this model by minimizing a mean dimension-wise cross-entropy loss over the elements of $p_{\theta}(x \mid u, a)$ and by minimizing a binary cross-entropy loss for the prediction $p_{\theta}(y \mid u, a)$. The MMD terms are computed by considering the $N$ samples $\{u_j \sim q_{\phi}(u_i \mid x_i, a_i)\}_{i=1}^N$ drawn during a mini-batch as samples from $q_{\phi}(U)$ \citep{ZhaoInfo}. The samples from $q_{\phi}(U \mid a)$ are taken as the subset of those drawn from the batch that correspond to $A = a$. As is suggested by \cite{ZhaoInfo}, we fix the $\lambda$ values in the loss such that each term is of the same order of magnitude. 


The architectural hyperparameters of this model are selected on the basis of minimization of the weighted loss on the validation set over one hundred iterations of random search and are selected separately for each combination of sensitive attribute and clinical outcome. The selected hyperparameters as shown in Table \ref{tab:params_vae}.

\subsection{Fair Predictor}
The fair predictor is trained with the objective given by equation \ref{eq:clp_loss}. The model architecture for the fair predictor is fixed to match that of the predictive component of the VAE, $p_{\theta}(y \mid u, a)$, and the weights are randomly intitialized. Since the training objectives requires counterfactual outcomes $y_{A\leftarrow a'}$, for each sample $\{x, y, a\}$ we randomly sample a single $u\sim q_\phi(u|x,y,a)$ and $y_{A \leftarrow a'} \sim p_{\theta}(y \mid u,a')$ for all $a' \neq a$ at both training and evaluation time.

In practice, we leverage a modified objective where the predictor $h_{\theta}$ takes $u$ and $a$ as input, rather than $x$ and $a$. Counterfactual predictions are then made on the basis of $h_{\theta}(u, a')$ rather than $h_{\theta}(x_{A \leftarrow a'}, a')$. For computing the CLP, rather than using the inverse sigmoid of the predicted probabilities of a positive outcome, we take the mean over the element-wise squared differences in the two-dimensional pre-softmax logits produced by the predictor. 

Relevant hyperparameters include $\lambda_{\mathrm{CLP}}$, $\lambda_{\mathrm{CF}}$, and an additional hyperparameter CF-Gradients which, when true, indicates whether gradients are propagated through the counterfactual samples in the CLP term of equation \ref{eq:clp_loss}. As previously indicated, we perform a grid search and select the model that minimizes the CLP component of the loss in equation $\ref{eq:clp_loss}$ for a fixed value of $\lambda_{\mathrm{CLP}}$ across the grid of other hyperparameters. The selected hyperparameters are shown in Table \ref{tab:params_final_classifier}.

\section{Hyperparameters} \label{appendix:params}
\begin{table}[htbp]
    \centering 
    \footnotesize
    \begin{tabular}{lrr}
    \toprule
      Parameter & Length of Stay $\geq$ 7 Days & Inpatient Mortality \\
    \midrule
     Batch Size &    512 &       512 \\
      Dropout Probability &    0.5 &      0.75 \\
     Hidden Dimension &    128 &       128 \\
     Learning Rate &  0.00001 &    0.0001 \\
     Layer Normalization &   True &      True \\
     Number of Hidden Layers &      3 &         2 \\
    \bottomrule
    \end{tabular}
\caption{Selected hyperparameters for the baseline predictor by outcome} \label{tab:params_baseline}
\end{table}

\begin{table}[htbp]
    \centering 
    \footnotesize
    \begin{tabular}{lrrrrrr}
    \toprule
    {} & \multicolumn{3}{c}{Length of Stay $\geq$ 7 Days} & \multicolumn{3}{c}{Inpatient Mortality} \\
    {} &     Age &  Gender & Race/Ethnicity &       Age &  Gender & Race/Ethnicity \\
    \midrule
    Batch Size            &     512 &     512 &      512 &       512 &     512 &      512 \\
    Dropout Probability VAE            &    0.25 &    0.25 &     0.75 &         0 &    0.25 &        0 \\
    Dropout Probability Predictor  &    0.25 &    0.25 &      0.5 &       0.5 &    0.25 &      0.5 \\
    Group Embedding Dimension       &      64 &      64 &       32 &        64 &      64 &       64 \\
    Hidden Dimension Predictor &     128 &     128 &      256 &       256 &     128 &      256 \\
    $\lambda_{\mathrm{y}}$ &      10 &      10 &       10 &        10 &      10 &       10 \\
    $\lambda_{\mathrm{MMD}}$            &   10000 &   10000 &    10000 &     10000 &   10000 &    10000 \\
    $\lambda_{\mathrm{MMD}_A}$      &    1000 &    1000 &     1000 &      1000 &    1000 &     1000 \\
    $\lambda_{\mathrm{x}}$ &    1000 &    1000 &     1000 &      1000 &    1000 &     1000 \\
    Latent Dimension            &     128 &     128 &      128 &       128 &     128 &      128 \\
    Learning Rate                    &  0.0001 &  0.0001 &    0.001 &     0.001 &  0.0001 &    0.001 \\
    Layer Normalization VAE             &   False &   False &     True &      True &   False &     True \\
    Layer Normalization Predictor  &    True &    True &     True &      True &    True &     True \\
    Number of Hidden Layers VAE            &       1 &       1 &        2 &         1 &       1 &        1 \\
    Number of Hidden Layers Predictor &       2 &       2 &        2 &         2 &       2 &        2 \\
    \bottomrule
    \end{tabular}
\caption{Selected hyperparameters for the VAE by outcome and sensitive attribute.} \label{tab:params_vae}
\end{table}

\begin{table}[htbp]
\footnotesize
\centering 
    \begin{tabular}{llllrr}
    \toprule
    Outcome & Sensitive Attribute & $\lambda_{CLP}$ &  CF-Gradients &  $\lambda_{CF}$ &  Learning Rate    \\
    \midrule
    Length of Stay $\geq$ 7 Days & Age & 0.00  &          True &                         0.1 &               0.0001 \\
              &          & 0.01  &          True &                         0.0 &               0.0100 \\
              &          & 0.10  &          True &                         0.0 &               0.0100 \\
              &          & 1.00  &          True &                         0.0 &               0.0100 \\
              &          & 10.00 &         False &                         0.0 &               0.0100 \\
              & Gender & 0.00  &         False &                         0.1 &               0.0100 \\
              &          & 0.01  &          True &                         0.1 &               0.0010 \\
              &          & 0.10  &          True &                         0.0 &               0.0010 \\
              &          & 1.00  &          True &                         0.0 &               0.0100 \\
              &          & 10.00 &          True &                         0.1 &               0.0010 \\
              & Race/Ethnicity & 0.00  &         False &                        10.0 &               0.0100 \\
              &          & 0.01  &          True &                         0.1 &               0.0100 \\
              &          & 0.10  &          True &                         0.0 &               0.0010 \\
              &          & 1.00  &          True &                         0.0 &               0.0010 \\
              &          & 10.00 &          True &                         1.0 &               0.0100 \\
    Inpatient Mortality & Age & 0.00  &         False &                         0.0 &               0.0100 \\
              &          & 0.01  &         False &                         0.0 &               0.0100 \\
              &          & 0.10  &          True &                         0.0 &               0.0010 \\
              &          & 1.00  &          True &                         0.1 &               0.0010 \\
              &          & 10.00 &          True &                         0.0 &               0.0010 \\
              & Gender & 0.00  &          True &                        10.0 &               0.0100 \\
              &          & 0.01  &          True &                         0.1 &               0.0100 \\
              &          & 0.10  &          True &                         0.0 &               0.0010 \\
              &          & 1.00  &          True &                         1.0 &               0.0100 \\
              &          & 10.00 &          True &                         0.1 &               0.0010 \\
              & Race/Ethnicity & 0.00  &          True &                        10.0 &               0.0100 \\
              &          & 0.01  &          True &                         1.0 &               0.0100 \\
              &          & 0.10  &          True &                         0.0 &               0.0100 \\
              &          & 1.00  &          True &                         0.0 &               0.0001 \\
              &          & 10.00 &          True &                         0.1 &               0.0010 \\
    \bottomrule
    \end{tabular}    
\caption{Selected hyperparameters for the models trained to be fair with respect to individual equalized counterfactual odds by outcome, sensitive attribute, and $\lambda_{\mathrm{CLP}}$.} \label{tab:params_final_classifier}
\end{table}

\clearpage
\section{Supplementary Tables} \label{appendix:tables}



\begin{table}[htbp]
    \centering 
    \begin{tabular}{lrrrrrrrr}
    \toprule
    {} & \multicolumn{4}{c}{Length of Stay $\geq 7$ Days} & \multicolumn{4}{c}{Inpatient Mortality} \\
    \cmidrule(lr){2-5} \cmidrule(lr){6-9}
    $\lambda_{CLP}$ & AUC-PRC & AUC-ROC & Brier &      CLP &   AUC-PRC & AUC-ROC &  Brier &      CLP \\
    \midrule
    N/A &   0.582 &   0.851 & 0.115 &      N/A &     0.267 &   0.893 & 0.0206 &      N/A \\
    0.0      &   0.563 &    0.84 & 0.118 &   0.0999 &     0.218 &   0.879 & 0.0207 &   0.0363 \\
    0.01     &   0.567 &   0.841 & 0.119 &   0.0335 &     0.208 &   0.868 & 0.0208 & 0.000355 \\
    0.1      &   0.568 &   0.842 & 0.118 &  0.00426 &     0.208 &   0.871 & 0.0207 & 7.12e-05 \\
    1.0      &    0.56 &   0.839 & 0.118 &    5e-05 &     0.203 &   0.872 & 0.0209 & 3.37e-06 \\
    10.0     &   0.558 &   0.835 & 0.134 & 9.82e-06 &    0.0772 &    0.76 & 0.0241 & 1.04e-07 \\
    \bottomrule
    \end{tabular}
    \caption{Model performance as a function of $\lambda_{\mathrm{CLP}}$ when \textbf{gender} is considered as a sensitive attribute. CLP is an aggregate measure of the extent to which a model satisfies individual equalized counterfactual odds and is computed as the mean per factual sample of the third term in equation \ref{eq:clp_loss}. N/A indicates the baseline model.} \label{tab:gender_performance}
\end{table}

\begin{table}[htbp]
    \centering 
    \begin{tabular}{lrrrrrrrr}
    \toprule
    {} & \multicolumn{4}{c}{Length of Stay $\geq 7$ Days} & \multicolumn{4}{c}{Inpatient Mortality} \\
    \cmidrule(lr){2-5} \cmidrule(lr){6-9}
    $\lambda_{CLP}$ & AUC-PRC & AUC-ROC & Brier &      CLP &   AUC-PRC & AUC-ROC &  Brier &      CLP \\
    \midrule
    N/A &   0.582 &   0.851 & 0.115 &      N/A &     0.267 &   0.893 & 0.0206 &      N/A \\
    0.0      &   0.542 &   0.822 & 0.125 &    0.107 &     0.203 &   0.869 &  0.024 &    0.223 \\
    0.01     &   0.555 &   0.836 & 0.122 &   0.0398 &     0.187 &   0.829 &  0.024 &   0.0191 \\
    0.1      &   0.555 &   0.836 & 0.122 &   0.0104 &     0.183 &   0.823 &  0.022 &   0.0055 \\
    1.0      &   0.555 &   0.836 & 0.119 &  0.00027 &     0.202 &   0.822 & 0.0225 & 0.000237 \\
    10.0     &   0.564 &   0.835 & 0.121 & 1.99e-05 &     0.162 &    0.81 & 0.0243 & 8.73e-07 \\
    \bottomrule
    \end{tabular}
    \caption{Model performance as a function of $\lambda_{\mathrm{CLP}}$ when \textbf{age} is considered as a sensitive attribute. CLP is an aggregate measure of the extent to which a model satisfies individual equalized counterfactual odds and is computed as the mean per factual sample of the third term in equation \ref{eq:clp_loss}. N/A indicates the baseline model.} \label{tab:age_performance}
\end{table}


\begin{table}[htbp]
    \centering 
    \begin{tabular}{llrrrrrr}
    \toprule
          \multicolumn{2}{c}{}   &    \multicolumn{6}{c}{$\lambda_{CLP}$}   \\
          \cmidrule(lr){3-8}
    Group & Metric &           N/A &    0.0 &   0.01 &    0.1 &    1.0 &   10.0        \\
    \midrule
    Female & AUC-PRC &     0.564 & 0.529 & 0.544 & 0.539 & 0.534 & 0.541 \\
         & AUC-ROC &     0.864 & 0.853 & 0.854 & 0.856 & 0.855 & 0.848 \\
         & Brier &    0.0993 & 0.102 & 0.104 & 0.103 & 0.101 & 0.116 \\
    Male & AUC-PRC &     0.597 & 0.587 &  0.59 & 0.593 & 0.584 & 0.589 \\
         & AUC-ROC &     0.829 & 0.815 & 0.818 & 0.817 & 0.822 &  0.82 \\
         & Brier &     0.136 &  0.14 & 0.141 &  0.14 & 0.138 & 0.155 \\
    \bottomrule
    \end{tabular}
    \caption{Model performance for prediction of \textbf{prolonged length of stay} on each group as a function of $\lambda_{\mathrm{CLP}}$ when \textbf{gender} is considered as a sensitive attribute. N/A indicates the baseline model.} \label{tab:los_gender_performance}
\end{table}

\begin{table}[htbp]
    \centering 
    \begin{tabular}{llrrrrrr}
    \toprule
          \multicolumn{2}{c}{}   &    \multicolumn{6}{c}{$\lambda_{CLP}$}   \\
          \cmidrule(lr){3-8}
    Group & Metric &           N/A &    0.0 &   0.01 &    0.1 &    1.0 &   10.0        \\
    \midrule
    \lbrack18, 30) & AUC-PRC &     0.608 &  0.548 &  0.581 &  0.597 &  0.611 &  0.597 \\
             & AUC-ROC &     0.885 &   0.84 &  0.868 &  0.869 &  0.875 &   0.87 \\
             & Brier &     0.098 &   0.11 &  0.106 &  0.104 & 0.0992 &  0.103 \\
    \lbrack30, 45) & AUC-PRC &     0.545 &  0.515 &  0.532 &  0.531 &  0.549 &  0.546 \\
             & AUC-ROC &     0.882 &  0.852 &  0.869 &  0.864 &  0.871 &  0.867 \\
             & Brier &     0.087 & 0.0925 & 0.0941 & 0.0923 & 0.0884 & 0.0895 \\
    \lbrack45, 65) & AUC-PRC &     0.606 &  0.554 &  0.562 &  0.575 &  0.579 &  0.591 \\
             & AUC-ROC &     0.849 &  0.816 &  0.834 &  0.838 &  0.839 &  0.839 \\
             & Brier &     0.123 &  0.135 &  0.133 &  0.129 &  0.126 &  0.129 \\
    \lbrack65, 89) & AUC-PRC &     0.564 &  0.537 &  0.525 &  0.534 &  0.533 &  0.556 \\
             & AUC-ROC &     0.817 &   0.79 &  0.803 &  0.802 &  0.804 &  0.807 \\
             & Brier &     0.131 &  0.142 &   0.14 &  0.139 &  0.137 &  0.136 \\
    \bottomrule
    \end{tabular}
    \caption{Model performance for prediction of \textbf{prolonged length of stay} on each group as a function of $\lambda_{\mathrm{CLP}}$ when \textbf{age} is considered as a sensitive attribute. N/A indicates the baseline model.} \label{tab:los_age_performance}
\end{table}

\begin{table}[htbp]
    \centering 
    \begin{tabular}{llrrrrrr}
    \toprule
          \multicolumn{2}{c}{}   &    \multicolumn{6}{c}{$\lambda_{CLP}$}   \\
          \cmidrule(lr){3-8}
    Group & Metric &           N/A &    0.0 &   0.01 &    0.1 &    1.0 &   10.0        \\
    \midrule
    Asian & AUC-PRC &     0.238 &  0.192 &  0.179 &  0.206 &  0.207 &  0.133 \\
          & AUC-ROC &       0.9 &  0.848 &  0.849 &  0.827 &  0.815 &  0.813 \\
          & Brier &    0.0217 & 0.0255 & 0.0254 & 0.0247 & 0.0237 & 0.0248 \\
    Black & AUC-PRC &     0.275 &  0.152 &  0.253 &  0.166 &  0.185 &  0.303 \\
          & AUC-ROC &     0.899 &  0.878 &  0.862 &  0.872 &   0.87 &   0.89 \\
          & Brier &    0.0153 & 0.0221 &  0.022 & 0.0244 & 0.0181 & 0.0185 \\
    Hispanic & AUC-PRC &     0.327 &  0.272 &  0.281 &   0.27 &  0.274 &  0.284 \\
          & AUC-ROC &     0.913 &  0.871 &  0.868 &  0.856 &  0.818 &  0.831 \\
          & Brier &    0.0202 & 0.0237 & 0.0228 & 0.0233 & 0.0219 & 0.0242 \\
    Other & AUC-PRC &     0.407 &  0.153 &  0.158 &  0.248 &  0.233 &  0.288 \\
          & AUC-ROC &     0.932 &  0.849 &  0.849 &  0.859 &  0.842 &  0.844 \\
          & Brier &    0.0137 & 0.0223 & 0.0206 & 0.0216 & 0.0171 &  0.018 \\
    Unknown & AUC-PRC &     0.683 &  0.603 &  0.596 &  0.514 &  0.572 &   0.55 \\
          & AUC-ROC &     0.964 &  0.947 &   0.95 &  0.919 &    0.9 &  0.898 \\
          & Brier &    0.0367 & 0.0481 & 0.0493 &  0.049 & 0.0425 & 0.0559 \\
    White & AUC-PRC &     0.183 &  0.136 &  0.143 &  0.137 &  0.137 &  0.135 \\
          & AUC-ROC &     0.869 &   0.84 &  0.837 &  0.791 &  0.764 &  0.768 \\
          & Brier &    0.0209 & 0.0259 & 0.0255 & 0.0257 &  0.023 & 0.0235 \\
    \bottomrule
    \end{tabular}
    \caption{Model performance for prediction of \textbf{inpatient mortality} on each group as a function of $\lambda_{\mathrm{CLP}}$ when \textbf{race/ethnicity} is considered as a sensitive attribute. N/A indicates the baseline model.} \label{tab:mortality_race_performance}
\end{table}

\begin{table}[htbp]
    \centering 
    \begin{tabular}{llrrrrrr}
    \toprule
          \multicolumn{2}{c}{}   &    \multicolumn{6}{c}{$\lambda_{CLP}$}   \\
          \cmidrule(lr){3-8}
    Group & Metric &           N/A &    0.0 &   0.01 &    0.1 &    1.0 &   10.0        \\
    \midrule
    Female & AUC-PRC &     0.289 &  0.235 &  0.215 &  0.201 &  0.223 & 0.0653 \\
         & AUC-ROC &     0.924 &   0.92 &  0.906 &  0.912 &  0.907 &  0.788 \\
         & Brier &     0.016 & 0.0159 & 0.0161 & 0.0163 & 0.0159 & 0.0194 \\
    Male & AUC-PRC &     0.255 &   0.23 &  0.216 &  0.231 &  0.205 & 0.0807 \\
         & AUC-ROC &     0.854 &   0.85 &  0.836 &  0.851 &  0.829 &  0.725 \\
         & Brier &    0.0264 & 0.0267 & 0.0267 & 0.0263 & 0.0268 & 0.0301 \\
    \bottomrule
    \end{tabular}
    \caption{Model performance for prediction of \textbf{inpatient mortality} on each group as a function of $\lambda_{\mathrm{CLP}}$ when \textbf{gender} is considered as a sensitive attribute. N/A indicates the baseline model.} \label{tab:mortality_gender_performance}
\end{table}

\begin{table}[htbp]
    \centering 
    \begin{tabular}{llrrrrrr}
    \toprule
          \multicolumn{2}{c}{}   &    \multicolumn{6}{c}{$\lambda_{CLP}$}   \\
          \cmidrule(lr){3-8}
    Group & Metric &           N/A &    0.0 &   0.01 &    0.1 &    1.0 &   10.0        \\
    \midrule
    \lbrack18, 30) & AUC-PRC &    0.0507 &  0.0589 &   0.052 &  0.0582 &  0.0516 &   0.023 \\
             & AUC-ROC &      0.83 &   0.807 &   0.642 &   0.675 &   0.629 &   0.836 \\
             & Brier &   0.00565 & 0.00684 & 0.00698 & 0.00606 & 0.00662 & 0.00831 \\
    \lbrack30, 45) & AUC-PRC &     0.333 &   0.241 &   0.208 &   0.242 &   0.236 &    0.21 \\
             & AUC-ROC &      0.97 &   0.907 &   0.943 &   0.912 &   0.907 &   0.883 \\
             & Brier &   0.00483 & 0.00502 & 0.00546 & 0.00505 & 0.00558 & 0.00833 \\
    \lbrack45, 65) & AUC-PRC &      0.33 &   0.199 &   0.194 &   0.207 &    0.21 &   0.179 \\
             & AUC-ROC &     0.906 &   0.874 &   0.881 &   0.861 &   0.876 &   0.853 \\
             & Brier &    0.0208 &  0.0266 &  0.0261 &  0.0228 &  0.0239 &  0.0254 \\
    \lbrack65, 89) & AUC-PRC &     0.258 &   0.223 &   0.219 &    0.22 &    0.23 &   0.212 \\
             & AUC-ROC &      0.84 &   0.813 &   0.802 &   0.799 &   0.804 &   0.795 \\
             & Brier &    0.0353 &  0.0404 &  0.0402 &   0.037 &  0.0386 &  0.0389 \\
    \bottomrule
    \end{tabular}
    \caption{Model performance for prediction of \textbf{inpatient mortality} on each group as a function of $\lambda_{\mathrm{CLP}}$ when \textbf{age} is considered as a sensitive attribute. N/A indicates the baseline model.} \label{tab:mortality_age_performance}
\end{table}

\clearpage
\section{Supplementary Figures} \label{appendix:figures}


\begin{figure}[htbp]
	\centering
	\includegraphics[width=0.9\linewidth]{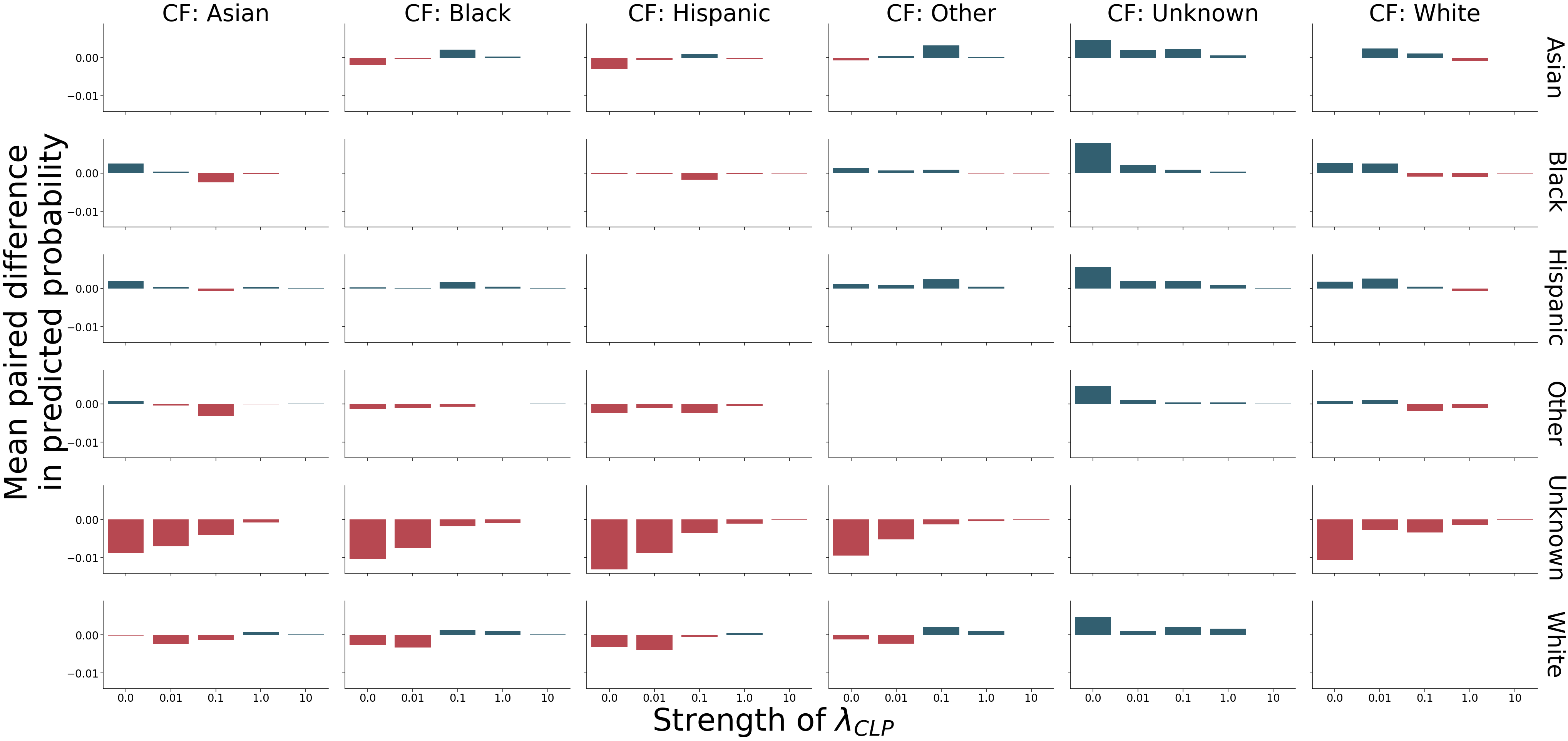}
	\caption{Mean difference in the counterfactual versus factual predicted probability of \textbf{inpatient mortality} conditioned on the outcome \textbf{not occurring} across \textbf{race/ethnicity} factual-counterfactual pairs. Positive values indicate a larger value for the counterfactual relative to the factual prediction.}
	\label{fig:bar_plot_mortality_race_0}
\end{figure}

\begin{figure}[htbp]
	\centering
	\includegraphics[width=0.9\linewidth]{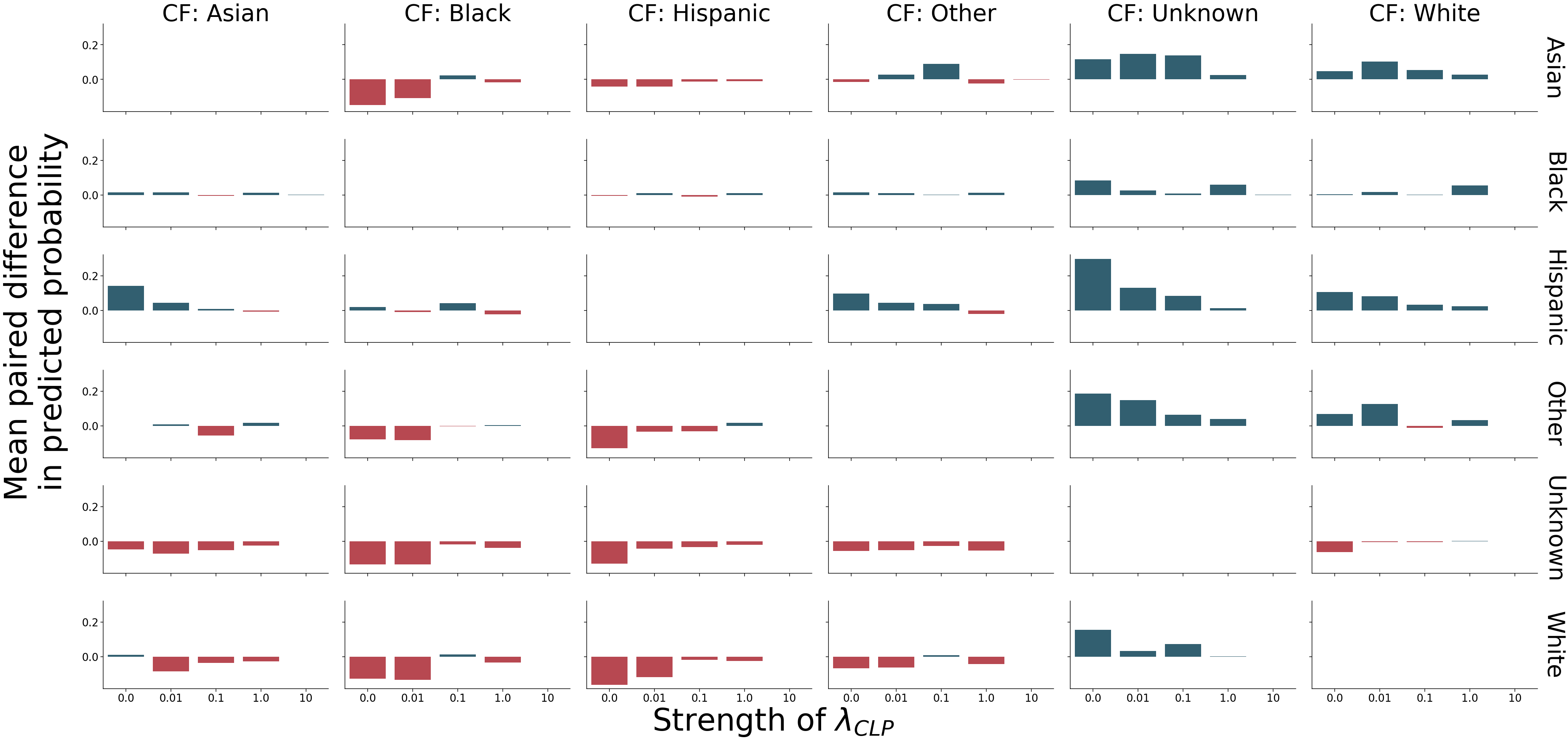}
	\caption{Mean difference in the counterfactual versus factual predicted probability of \textbf{inpatient mortality} conditioned on the outcome \textbf{occurring} across \textbf{race/ethnicity} factual-counterfactual pairs. Positive values indicate a larger value for the counterfactual relative to the factual prediction.}
	\label{fig:bar_plot_mortality_race_1}
\end{figure}


\begin{figure}[htbp]
	\centering
	\includegraphics[width=0.99\linewidth]{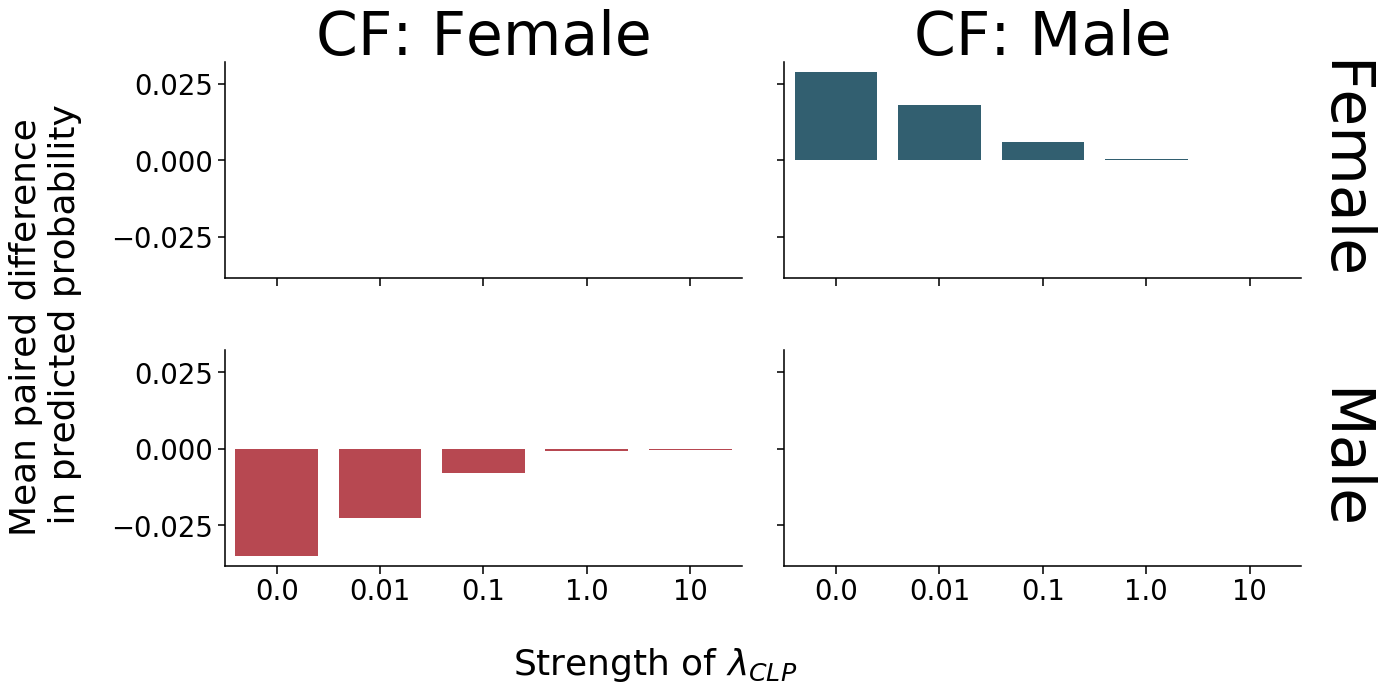}
	\caption{Mean difference in the counterfactual versus factual predicted probability of a \textbf{length of stay} greater than or equal to seven days conditioned on the outcome \textbf{not occurring} across \textbf{gender} factual-counterfactual pairs. Positive values indicate a larger value for the counterfactual relative to the factual prediction.}
	\label{fig:bar_plot_los_gender_0}
\end{figure}

\begin{figure}[htbp]
	\centering
	\includegraphics[width=0.99\linewidth]{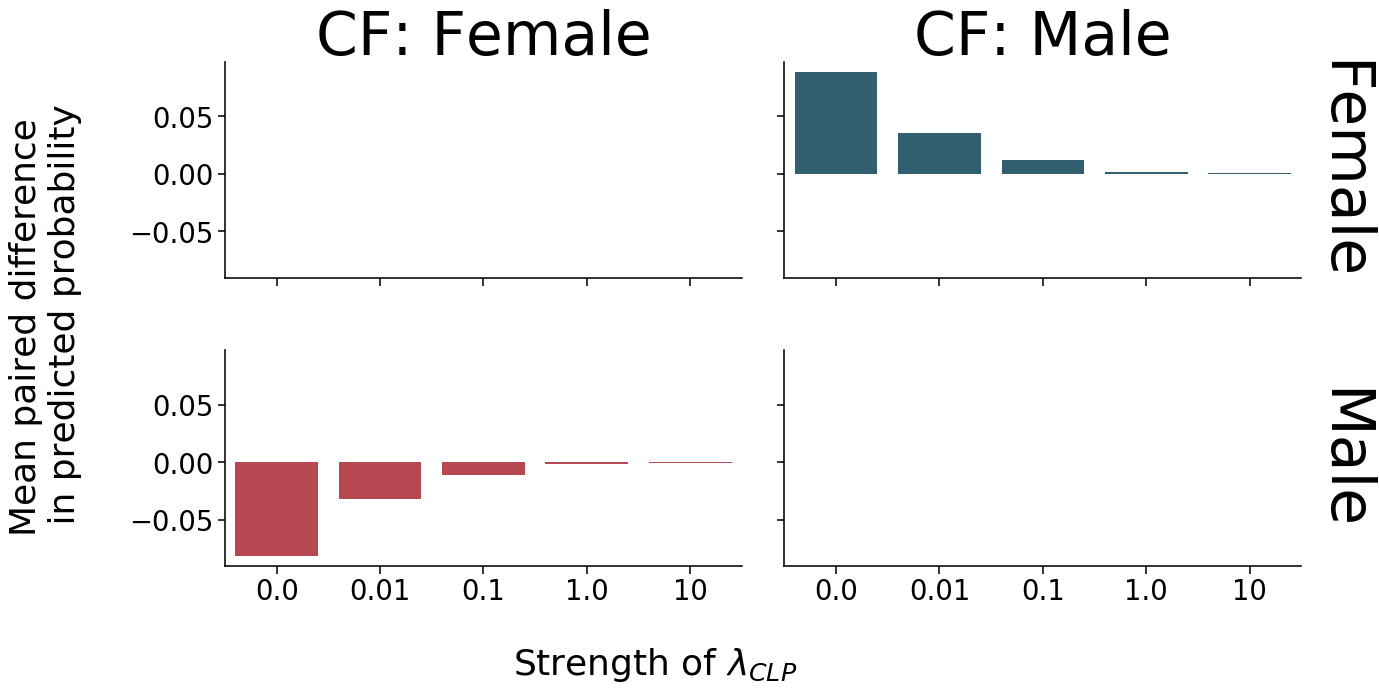}
    \caption{Mean difference in the counterfactual versus factual predicted probability of a \textbf{length of stay} greater than or equal to seven days conditioned on the outcome \textbf{occurring} across \textbf{gender} factual-counterfactual pairs. Positive values indicate a larger value for the counterfactual relative to the factual prediction.}
	\label{fig:bar_plot_los_gender_1}
\end{figure}

\begin{figure}[htbp]
	\centering
	\includegraphics[width=0.99\linewidth]{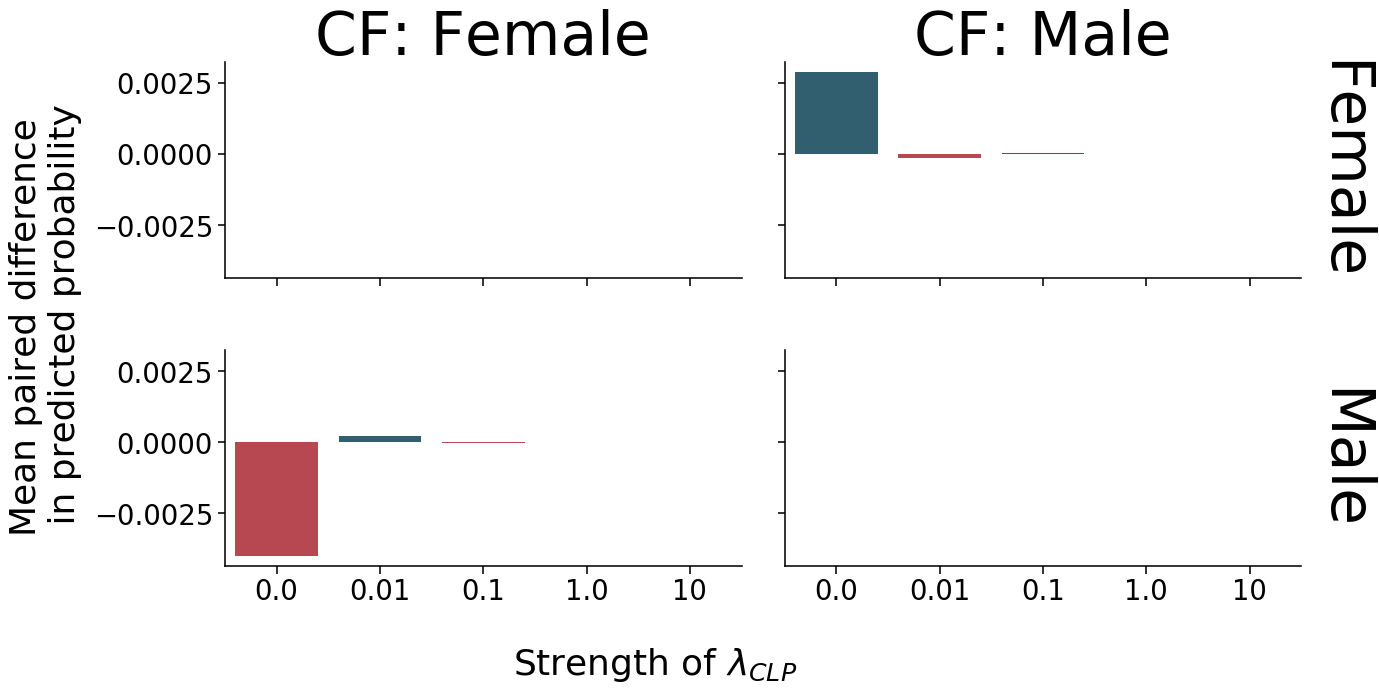}
	\caption{Mean difference in the counterfactual versus factual predicted probability of \textbf{inpatient mortality} conditioned on the outcome \textbf{not occurring} across \textbf{gender} factual-counterfactual pairs. Positive values indicate a larger value for the counterfactual relative to the factual prediction.}
	\label{fig:bar_plot_mortality_gender_0}
\end{figure}

\begin{figure}[htbp]
	\centering
	\includegraphics[width=0.99\linewidth]{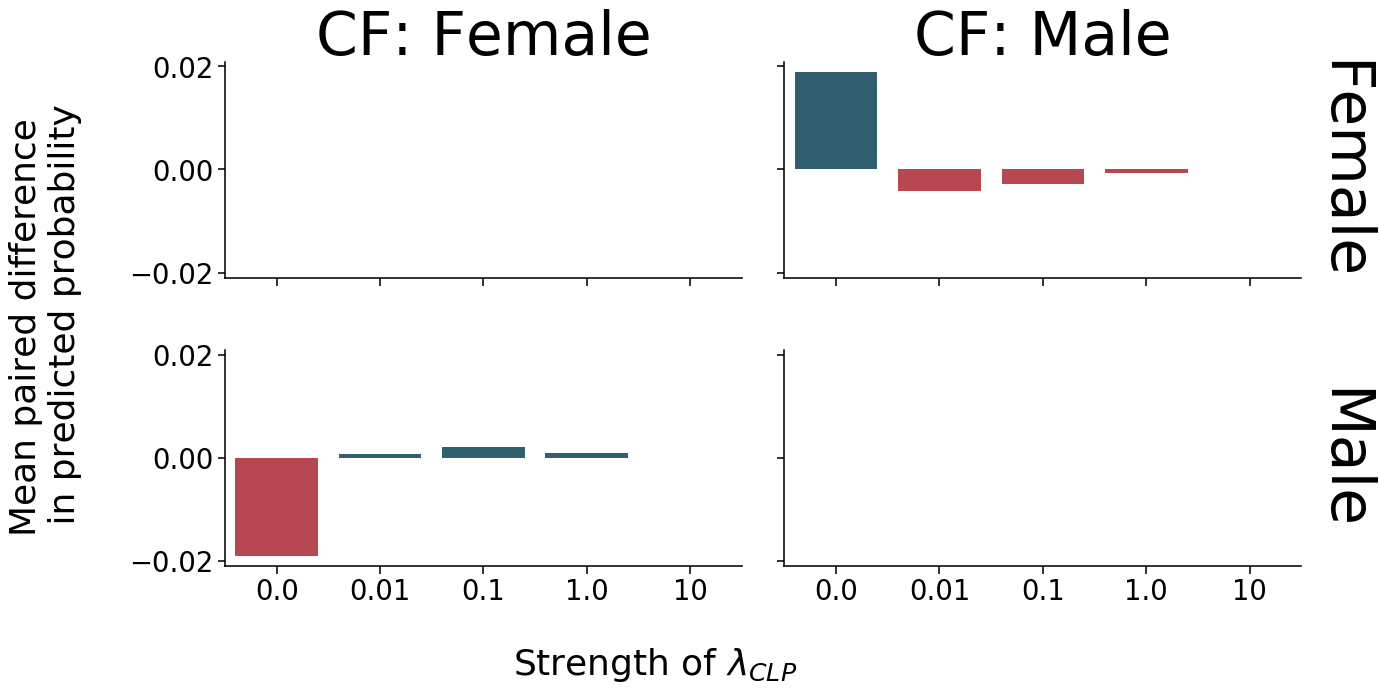}
	\caption{Mean difference in the counterfactual versus factual predicted probability of \textbf{inpatient mortality} conditioned on the outcome \textbf{occurring} across \textbf{gender} factual-counterfactual pairs. Positive values indicate a larger value for the counterfactual relative to the factual prediction.}
	\label{fig:bar_plot_mortality_gender_1}
\end{figure}


\begin{figure}[htbp]
	\centering
	\includegraphics[width=0.99\linewidth]{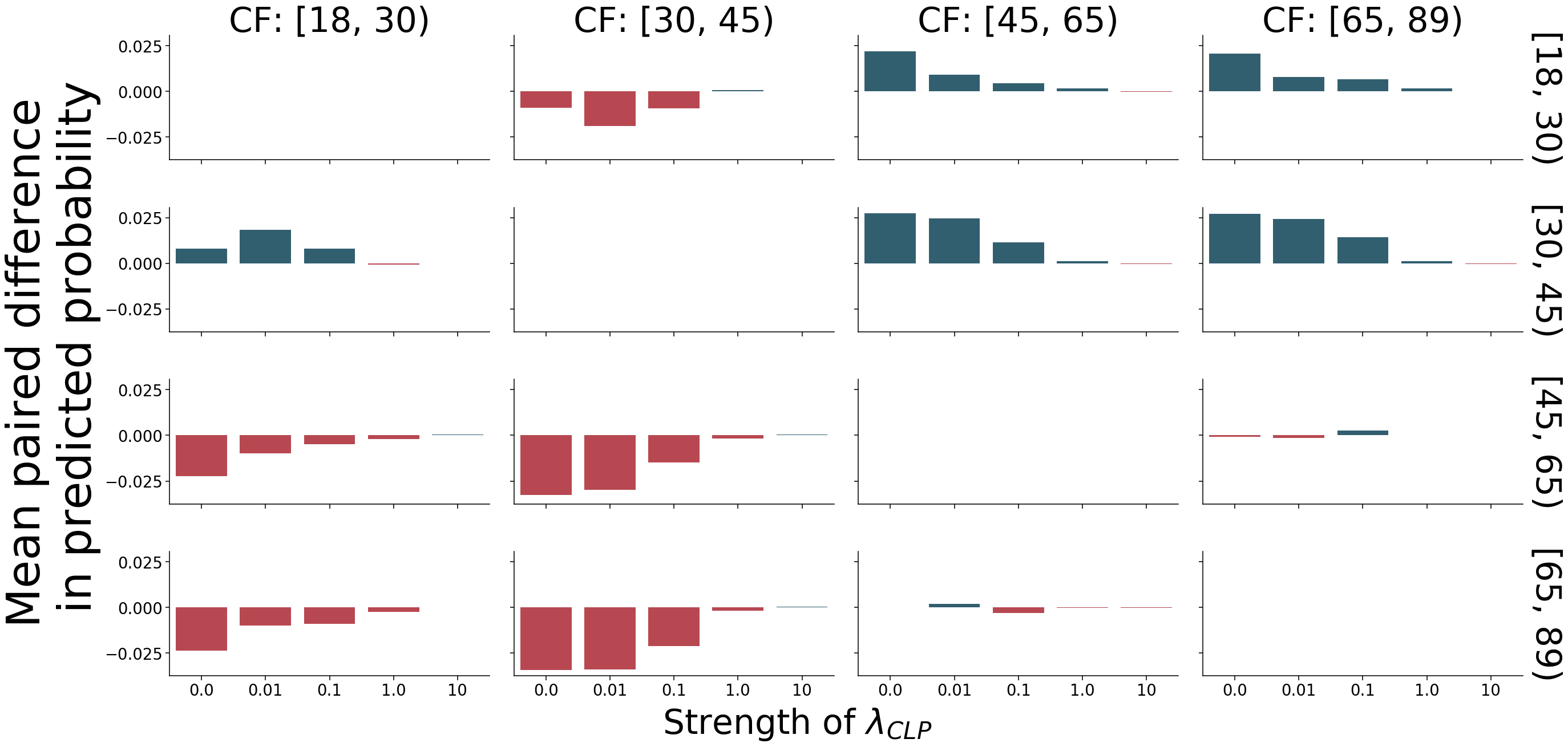}
	\caption{Mean difference in the counterfactual versus factual predicted probability of a \textbf{length of stay} greater than or equal to seven days conditioned on the outcome \textbf{not occurring} across \textbf{age} factual-counterfactual pairs. Positive values indicate a larger value for the counterfactual relative to the factual prediction.}
	\label{fig:bar_plot_los_age_0}
\end{figure}

\begin{figure}[htbp]
	\centering
	\includegraphics[width=0.99\linewidth]{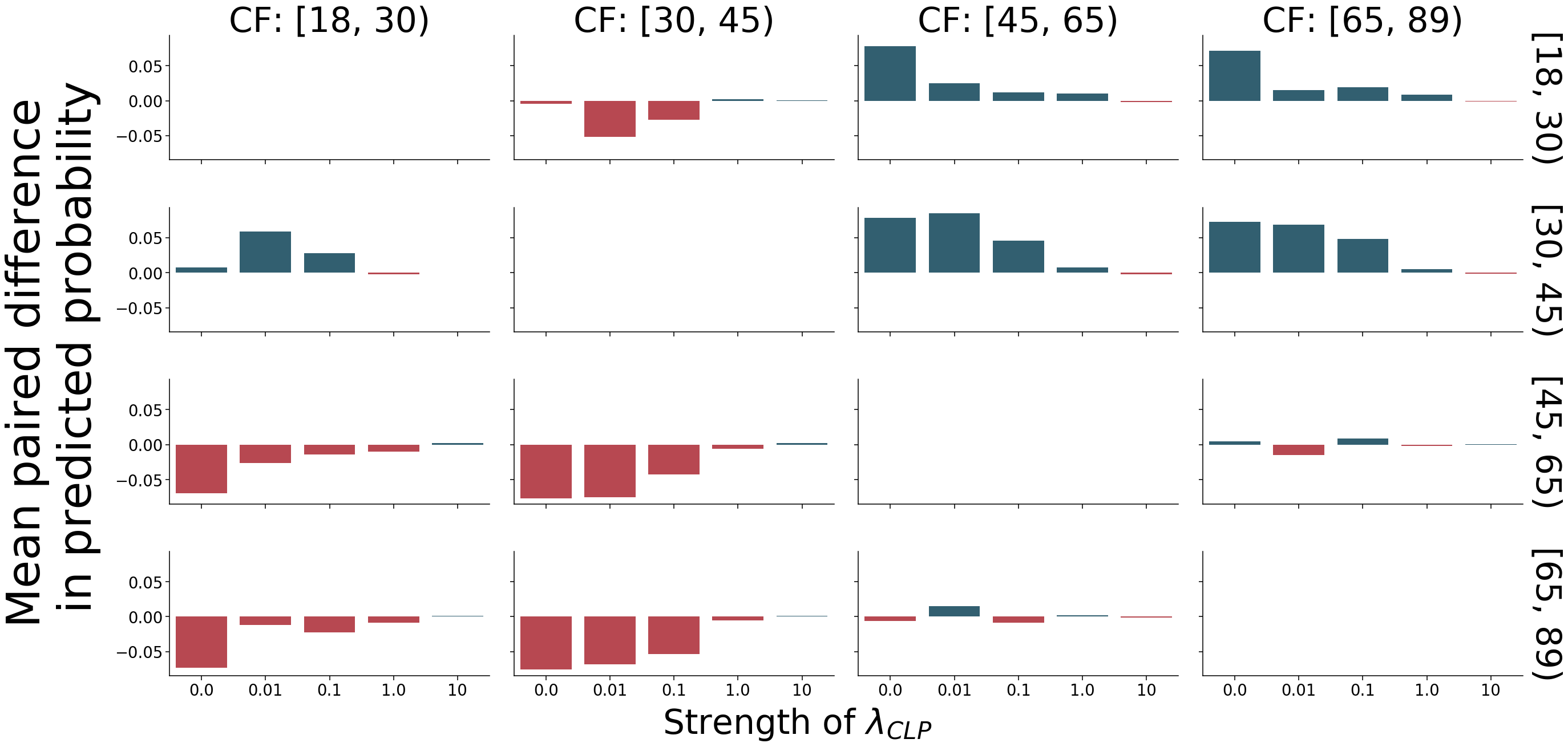}
    \caption{Mean difference in the counterfactual versus factual predicted probability of a \textbf{length of stay} greater than or equal to seven days conditioned on the outcome \textbf{occurring} across \textbf{age} factual-counterfactual pairs. Positive values indicate a larger value for the counterfactual relative to the factual prediction.}
	\label{fig:bar_plot_los_age_1}
\end{figure}

\begin{figure}[htbp]
	\centering
	\includegraphics[width=0.99\linewidth]{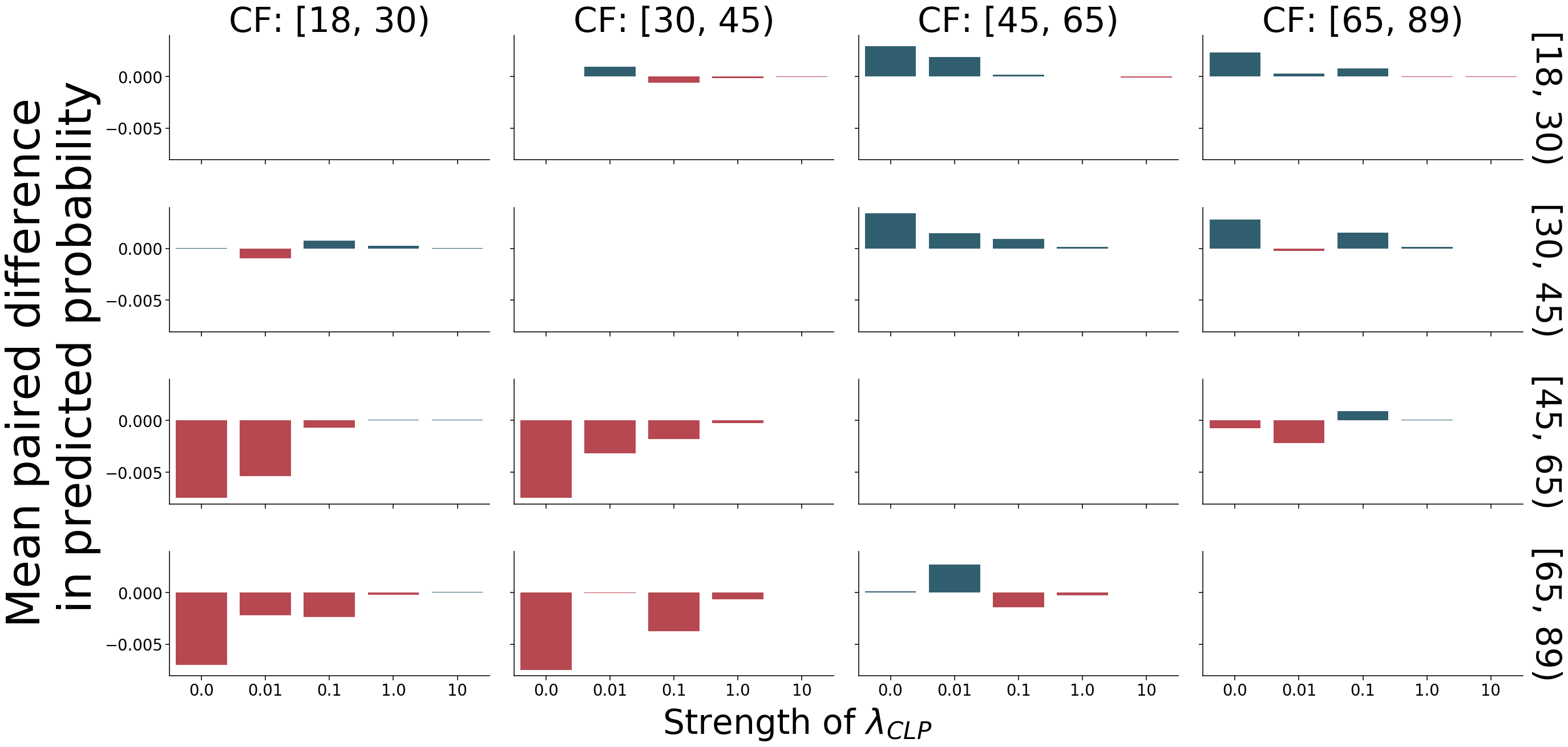}
	\caption{Mean difference in the counterfactual versus factual predicted probability of \textbf{inpatient mortality} conditioned on the outcome \textbf{not occurring} across \textbf{age} factual-counterfactual pairs. Positive values indicate a larger value for the counterfactual relative to the factual prediction.}
	\label{fig:bar_plot_mortality_age_0}
\end{figure}

\begin{figure}[htbp]
	\centering
	\includegraphics[width=0.99\linewidth]{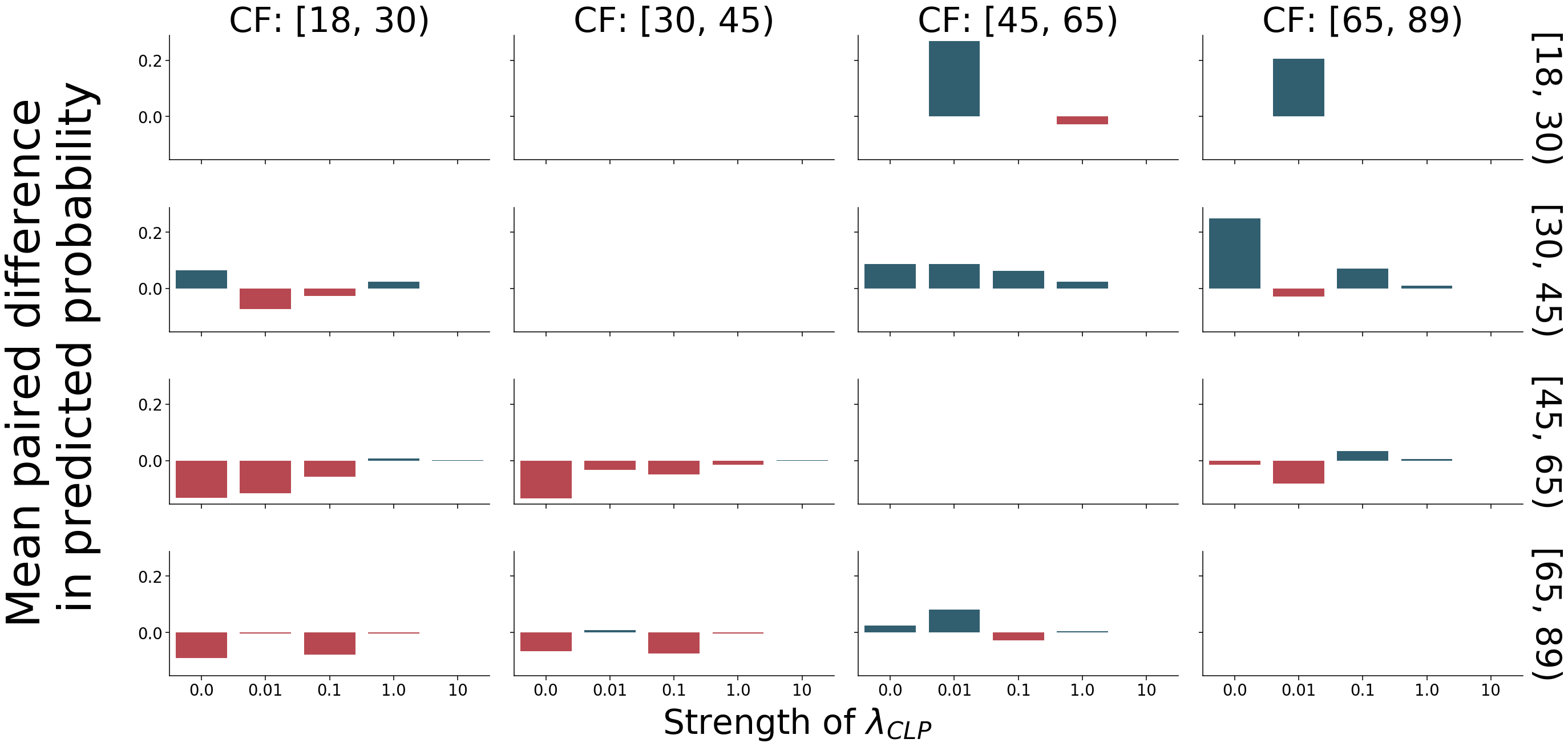}
	\caption{Mean difference in the counterfactual versus factual predicted probability of \textbf{inpatient mortality} conditioned on the outcome \textbf{occurring} across \textbf{age} factual-counterfactual pairs. Positive values indicate a larger value for the counterfactual relative to the factual prediction.}
	\label{fig:bar_plot_mortality_age_1}
\end{figure}

\end{document}